\documentclass[10pt,twocolumn,letterpaper]{article}

\usepackage{wacv}
\usepackage{times}
\usepackage{epsfig}
\usepackage{graphicx}
\usepackage{amsmath}
\usepackage{amssymb}
\usepackage{booktabs}
\usepackage[dvipsnames]{xcolor}         
\usepackage{tabularx}
\usepackage{etoolbox}
\usepackage{xspace}
\usepackage{multirow}
\usepackage{xspace}
\usepackage{ulem}
\usepackage{array}
\usepackage{pifont}
\usepackage{colortbl}
\usepackage{mathtools}
\usepackage[flushleft]{threeparttable}

\newcommand{\xmark}{\ding{55}}%
\definecolor{customgreen}{RGB}{34,139,34}  

\usepackage{xspace}
\usepackage[export]{adjustbox}

\usepackage{bold-extra}

\makeatletter

\definecolor{ubpubColor}{rgb}{0.43, 0.5, 0.5}
\definecolor{backboneColor}{rgb}{0.423, 0.325, 0.365}
\definecolor{fpnColor}{rgb}{0.255, 0.498, 0.416}

\newcommand{\PAR}[1]{\vskip3pt \noindent {\bf #1~}}

\newif\ifCommentsEnabled

\CommentsEnabledtrue

\ifCommentsEnabled

\newcommand{\bastian}[1]{\textcolor{red}{\textbf{Bastian: }{#1}}}
\newcommand{\achal}[1]{\textcolor{magenta}{\textbf{Achal: }{#1}}}
\newcommand{\tarasha}[1]{\textcolor{orange}{\textbf{Tarasha: }{#1}}}
\newcommand{\jono}[1]{\textcolor{blue}{\textbf{Jono: }{#1}}}
\newcommand{\paul}[1]{\textcolor{green}{\textbf{Paul: }{#1}}}
\newcommand{\ali}[1]{\textcolor{brown}{\textbf{Ali: }{#1}}}
\newcommand{\deva}[1]{\textcolor{red}{\textbf{Deva: }{#1}}}
\newcommand{\todo}[1]{\textcolor{red}{\small Todo:\,#1}\PackageWarning{TODO:}{#1!}}
\newcommand{\TODO}[1]{\textcolor{red}{#1}}

\else

\newcommand{\bastian}[1]{}
\newcommand{\achal}[1]{}
\newcommand{\tarasha}[1]{}
\newcommand{\jono}[1]{}
\newcommand{\paul}[1]{}
\newcommand{\ali}[1]{}
\newcommand{\deva}[1]{}
\newcommand{\todo}[1]{}
\newcommand{\TODO}[1]{}

\fi

\newcolumntype{P}[1]{>{\centering\arraybackslash}p{#1}}

\usepackage{pifont}

\newcommand{\JnF}{\mathcal{J}\&\mathcal{F}}

\newcommand{\hotadet}[1]{$\mathrm{HOTA}_\text{det}$}
\newcommand{\hotaobj}[1]{$\mathrm{HOTA}^\text{obj}$}
\newcommand{\detA}[1]{$\mathrm{DetA}$}
\newcommand{\assA}[1]{$\mathrm{AssA}$}
\newcommand{\detRe}[1]{$\mathrm{DetRe}$}
\newcommand{\owta}[1]{$\mathrm{OWTA}$}
\newcommand{\hotacls}[1]{$\mathrm{HOTA}_\text{cls}$}

\newif\ifArxivMode

\def\wacvPaperID{1536} 

\wacvalgorithmstrack   

\wacvfinalcopy 

\ifwacvfinal
\usepackage[breaklinks=true,bookmarks=false]{hyperref}
\else
\usepackage[pagebackref=true,breaklinks=true,colorlinks,bookmarks=false]{hyperref}
\fi

\def\DatasetAbbrev{BURST}
\def\hota{$\mathrm{HOTA}$}
\def\hotacommon{$\mathrm{HOTA}_\text{com}$}
\def\hotauncommon{$\mathrm{HOTA}_\text{unc}$}
\def\hotaall{$\mathrm{HOTA}_\text{all}$}
\def\numclasses{482}

\begin{document}

\title{BURST: A Benchmark for Unifying Object Recognition, Segmentation and Tracking in Video}

\newcommand{\titlespace}[0]{\quad}
\author{
Ali Athar$^1$ %
\titlespace
Jonathon Luiten$^{1,2}$
\titlespace
Paul Voigtlaender$^3$%
\titlespace
Tarasha Khurana$^2$%
\titlespace
Achal Dave$^4$
\\[2pt]
Bastian Leibe$^1$
\titlespace
Deva Ramanan$^2$\\[5pt]
$^1$ RWTH Aachen University, Germany 
\quad 
$^2$ Carnegie Mellon University, USA
\quad
$^3$ Google
\quad
$^4$ Amazon
\\[5pt]
{\tt\small \{athar,luiten,leibe\}@vision.rwth-aachen.de}
\quad {\tt\small \{tkhurana,deva\}@cs.cmu.edu} \\
{\tt\small voigtlaender@google.com \quad achald@amazon.com}
}

\maketitle

\ArxivModetrue

\begin{abstract}
Multiple existing benchmarks involve tracking and segmenting objects in video e.g., Video Object Segmentation (VOS) and Multi-Object Tracking and Segmentation (MOTS), but there is little interaction between them due to the use of disparate benchmark datasets and metrics (e.g. $\JnF$, mAP, sMOTSA).  
As a result, published works usually target a particular benchmark, and are not easily comparable to each another.
We believe that the development of generalized methods that can tackle multiple tasks requires greater cohesion among these research sub-communities.
In this paper, we aim to facilitate this by proposing \DatasetAbbrev{}, a dataset which contains thousands of diverse videos with high-quality object masks, and an associated benchmark with six tasks involving object tracking and segmentation in video.
All tasks are evaluated using the same data and comparable metrics, which
enables researchers to consider them in unison, and hence, more effectively pool knowledge from different methods across different tasks.
Additionally, we demonstrate several baselines for all tasks and show that approaches for one task can be applied to another with a quantifiable and explainable performance difference. 
Dataset annotations and evaluation code is available at: \url{https://github.com/Ali2500/BURST-benchmark}.
\end{abstract}

\section{Introduction}
\label{sec:intro}

\begin{figure}[t]
  \centering
  \setlength\tabcolsep{1pt}
  \renewcommand{\arraystretch}{0.5}
  \def\cellWidth{0.32\linewidth}
  \begin{tabularx}{\textwidth}{ccc}%
       \includegraphics[width=\cellWidth]{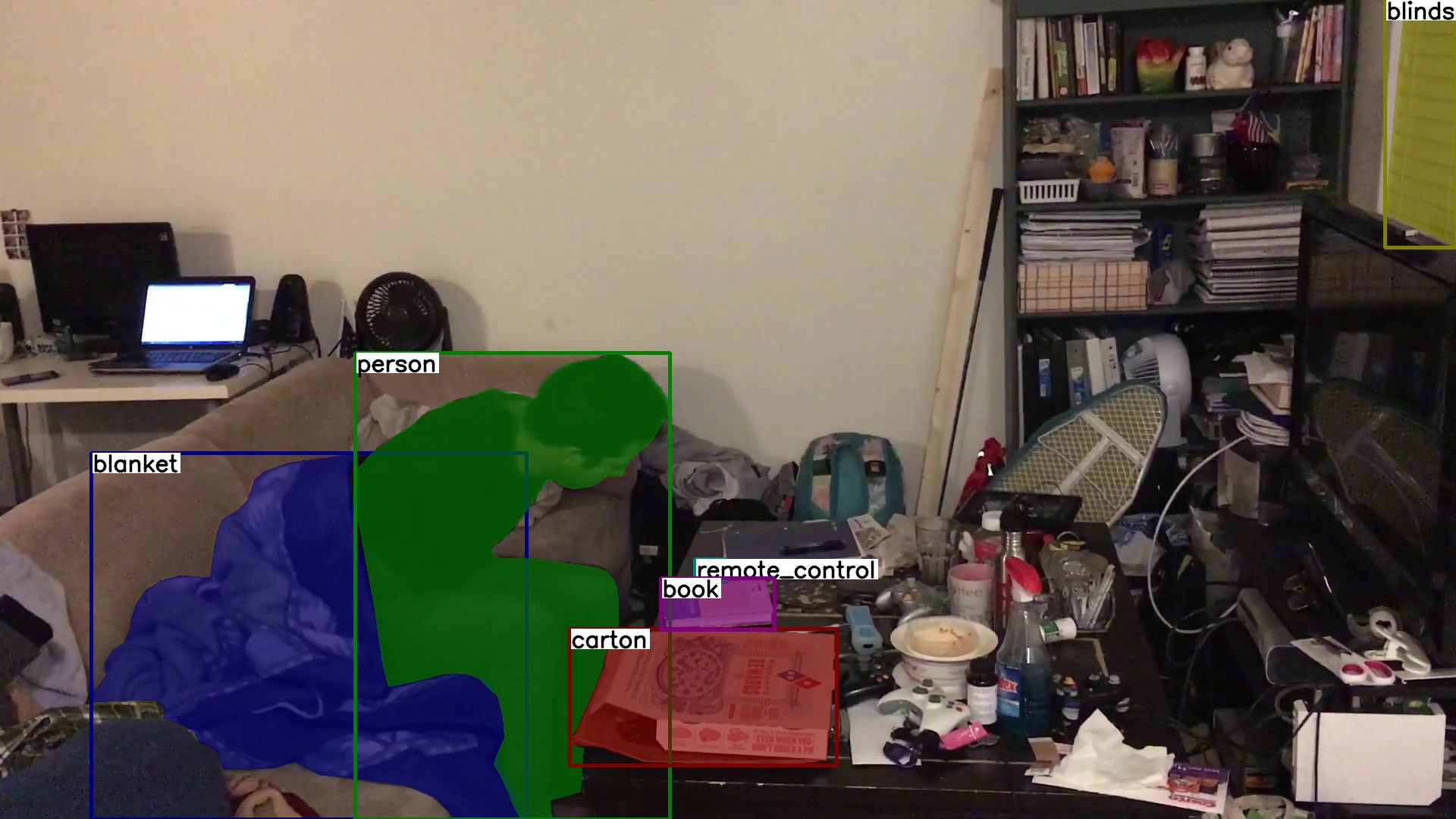}&%
       \includegraphics[width=\cellWidth]{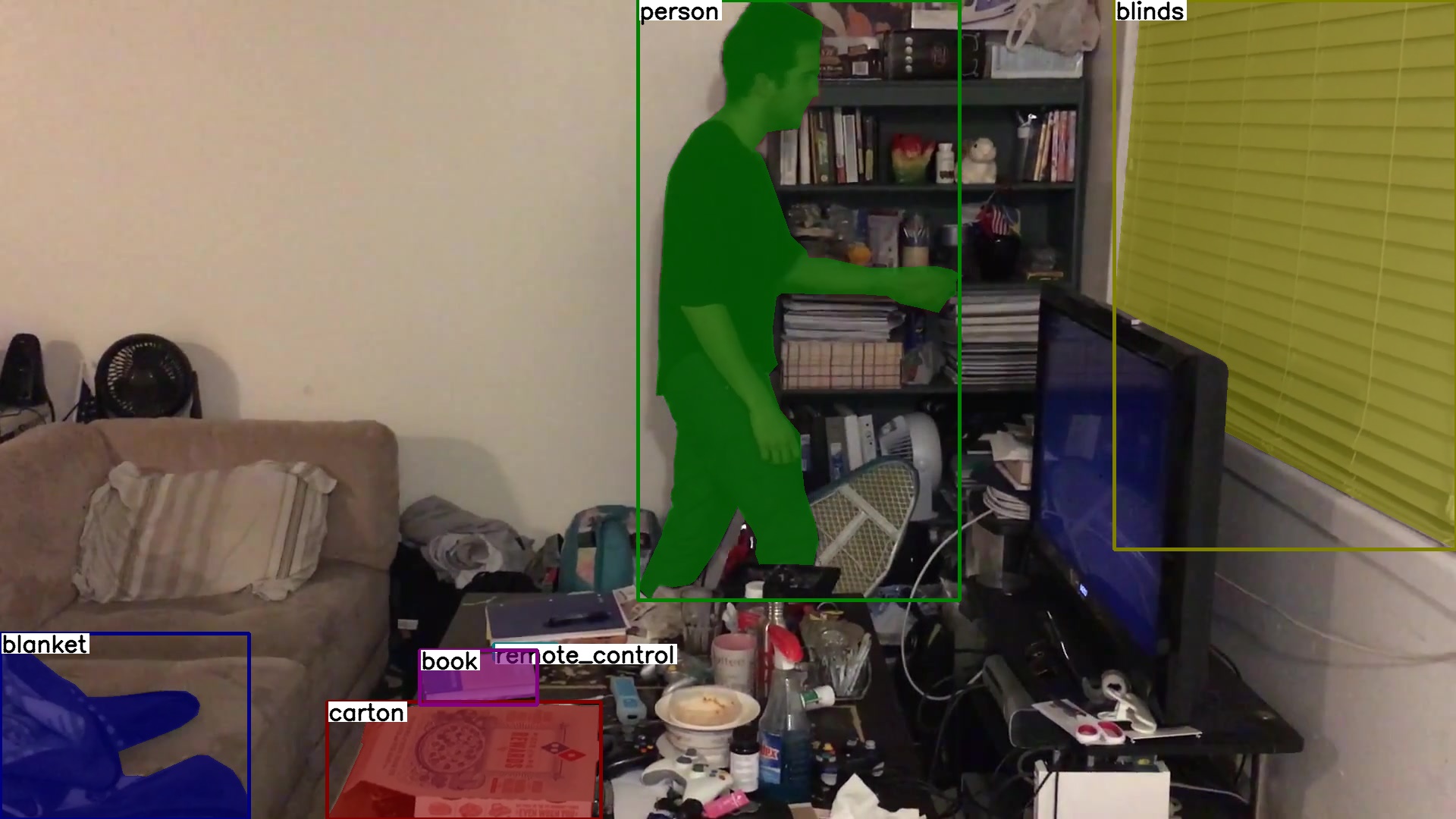}&%
       \includegraphics[width=\cellWidth]{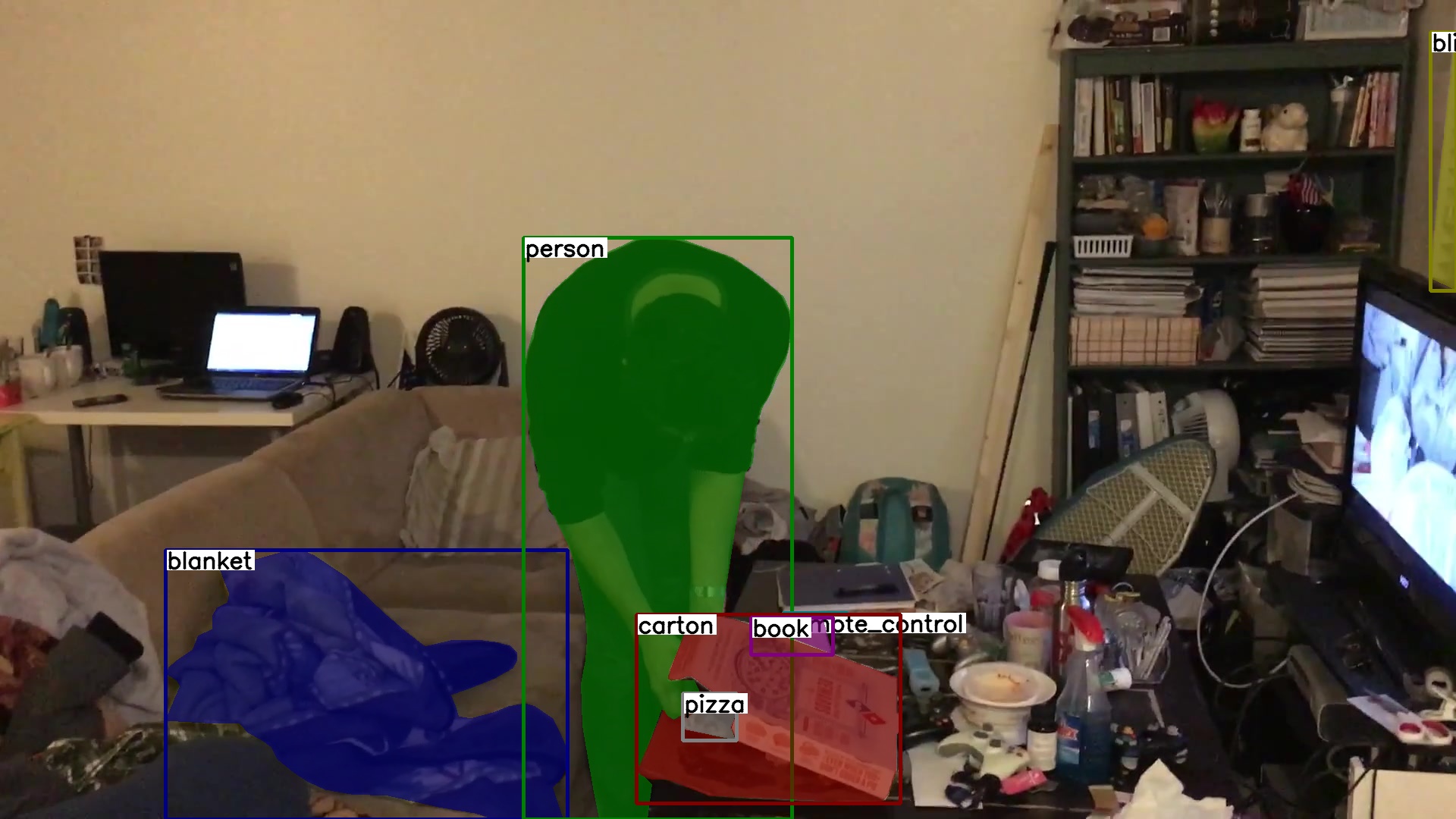}
       \\
       \includegraphics[width=\cellWidth]{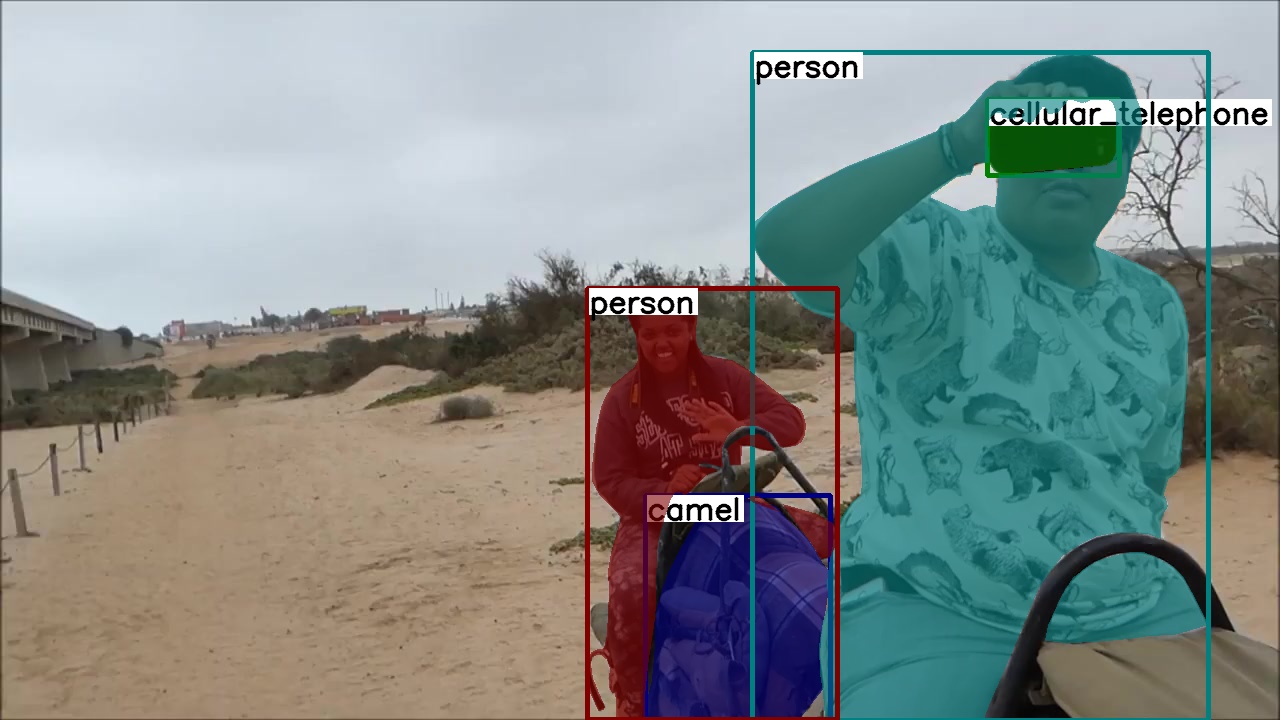}&%
       \includegraphics[width=\cellWidth]{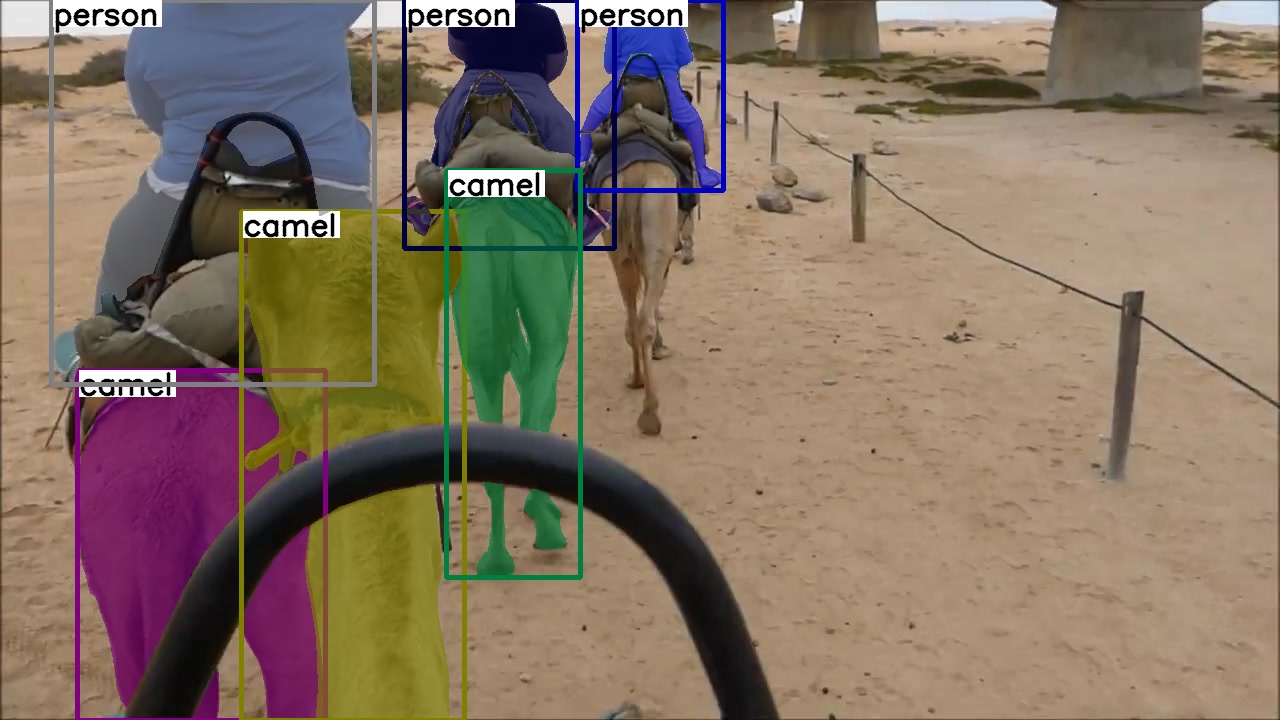}&%
       \includegraphics[width=\cellWidth]{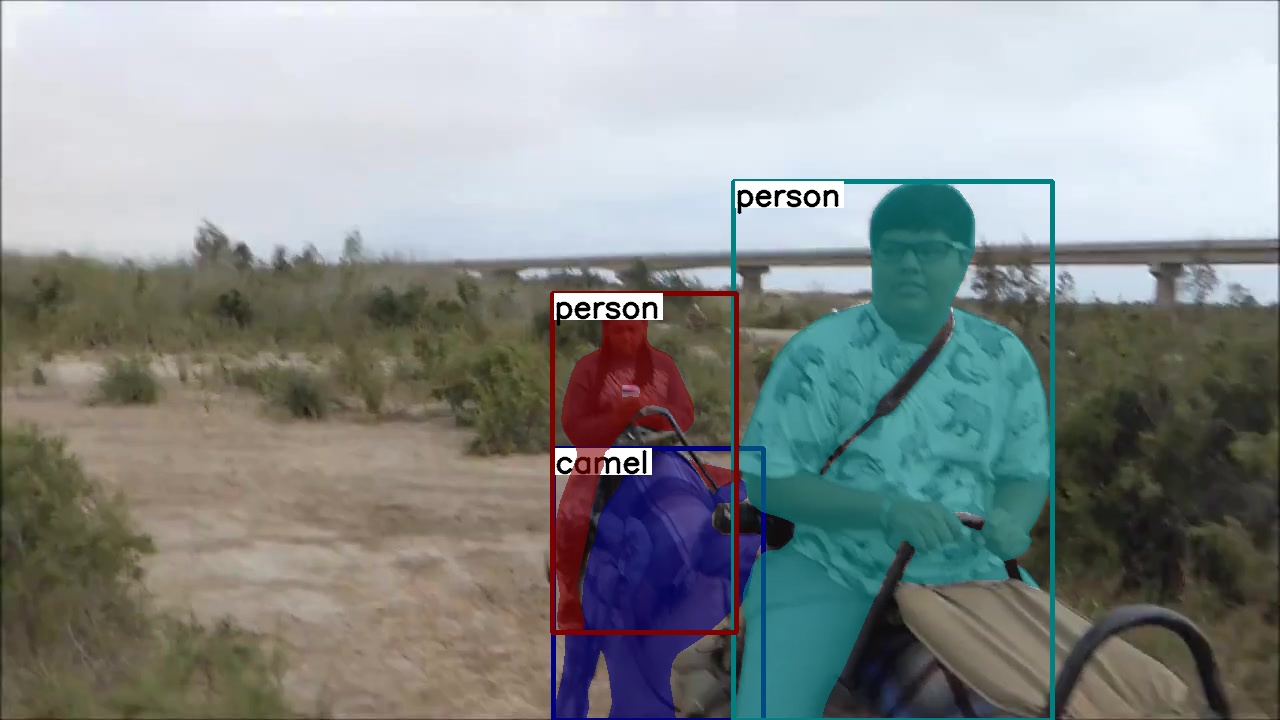}
       \\
       \includegraphics[width=\cellWidth]{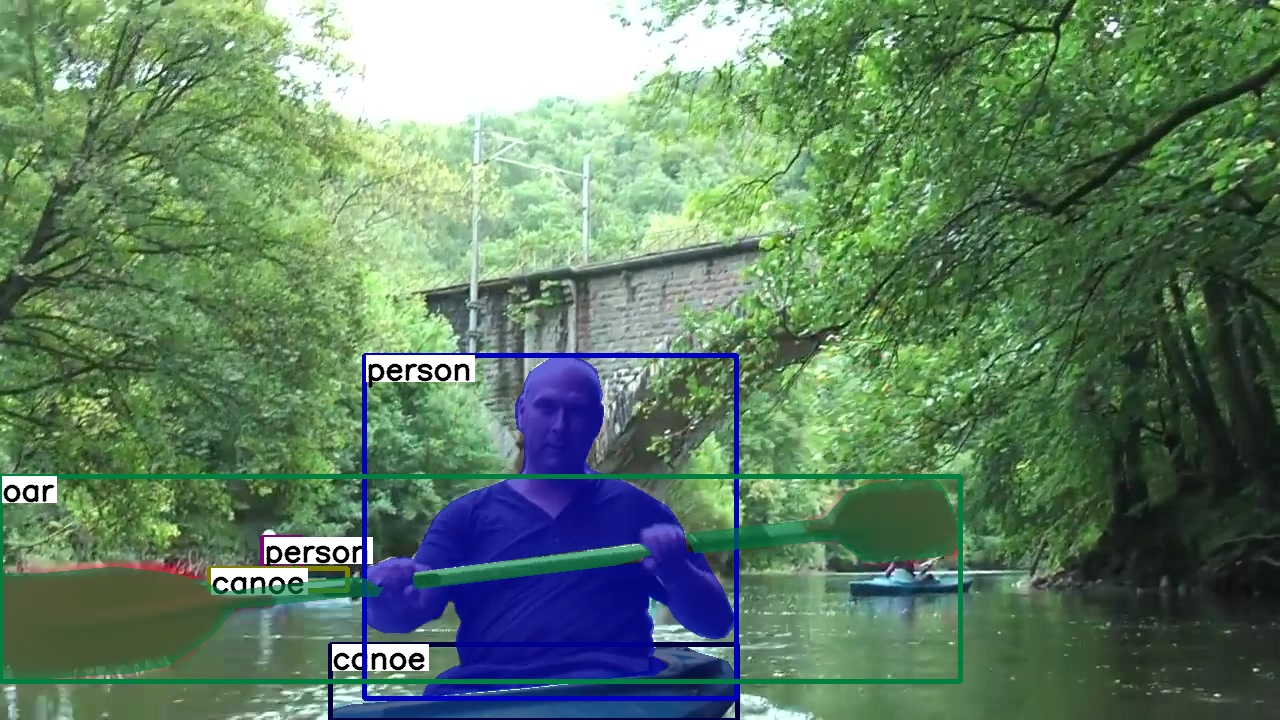}&%
       \includegraphics[width=\cellWidth]{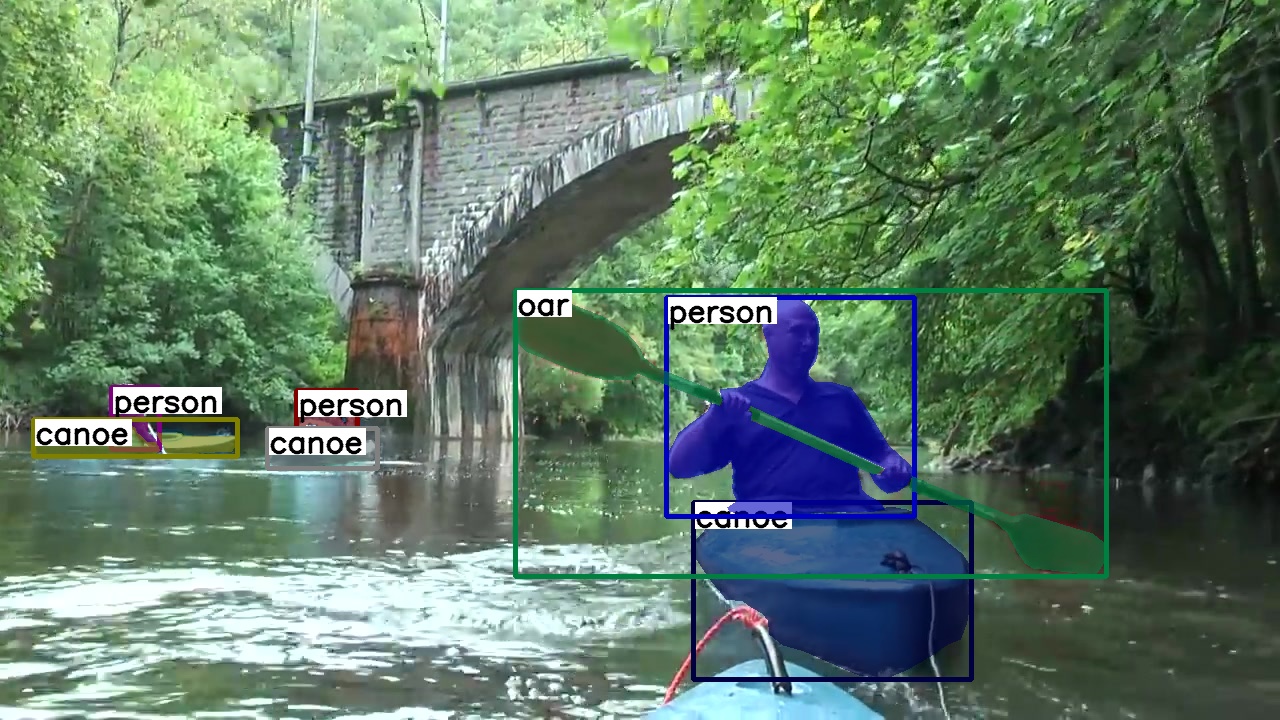}&%
       \includegraphics[width=\cellWidth]{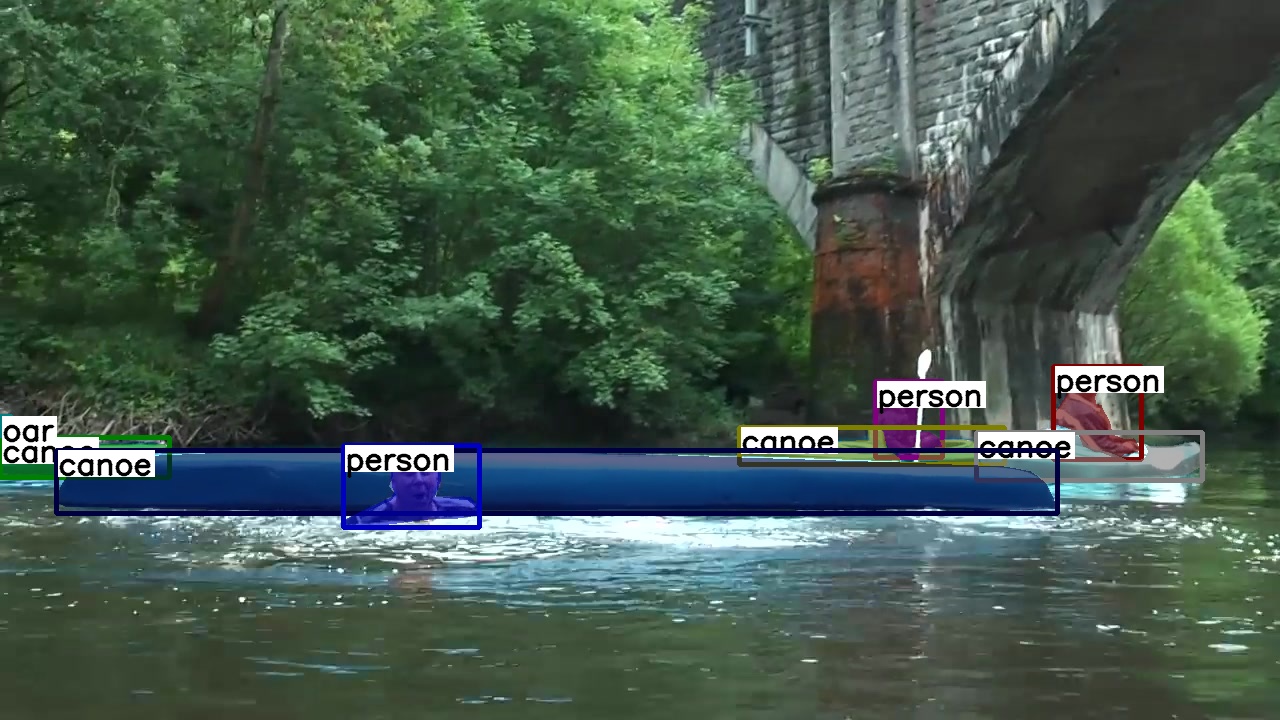}%
       \\
       \includegraphics[width=\cellWidth]{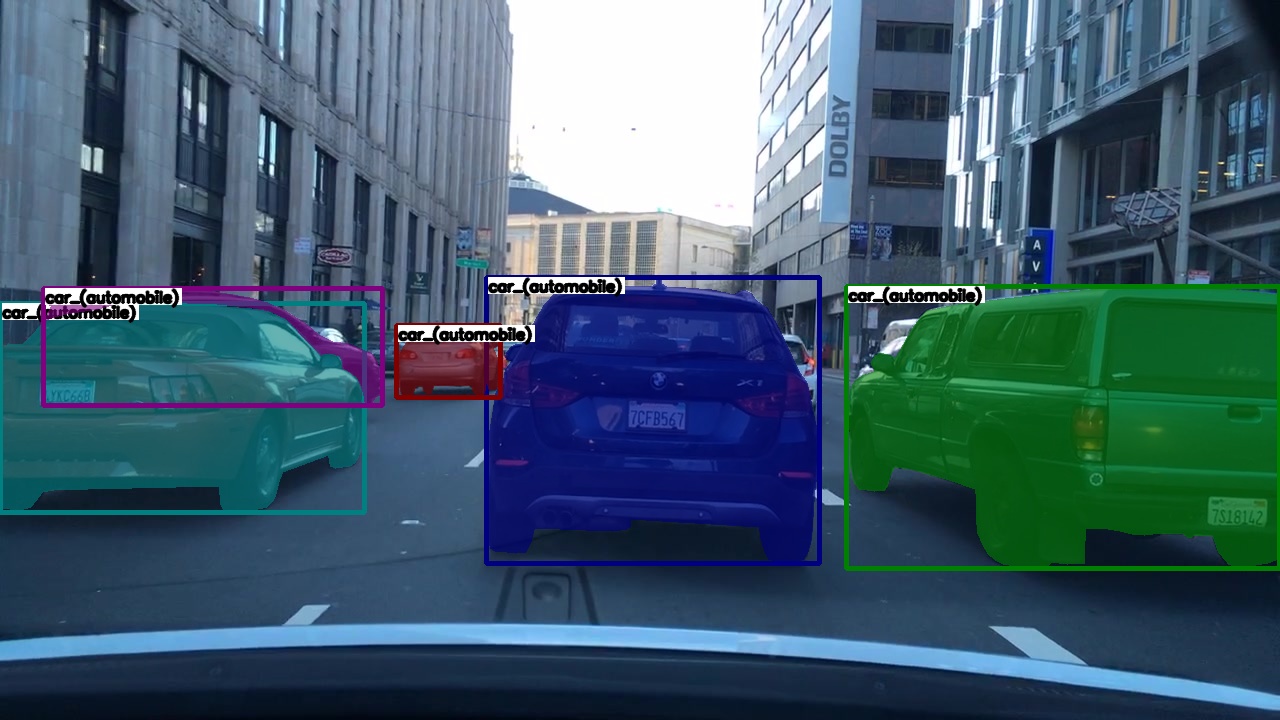}&%
       \includegraphics[width=\cellWidth]{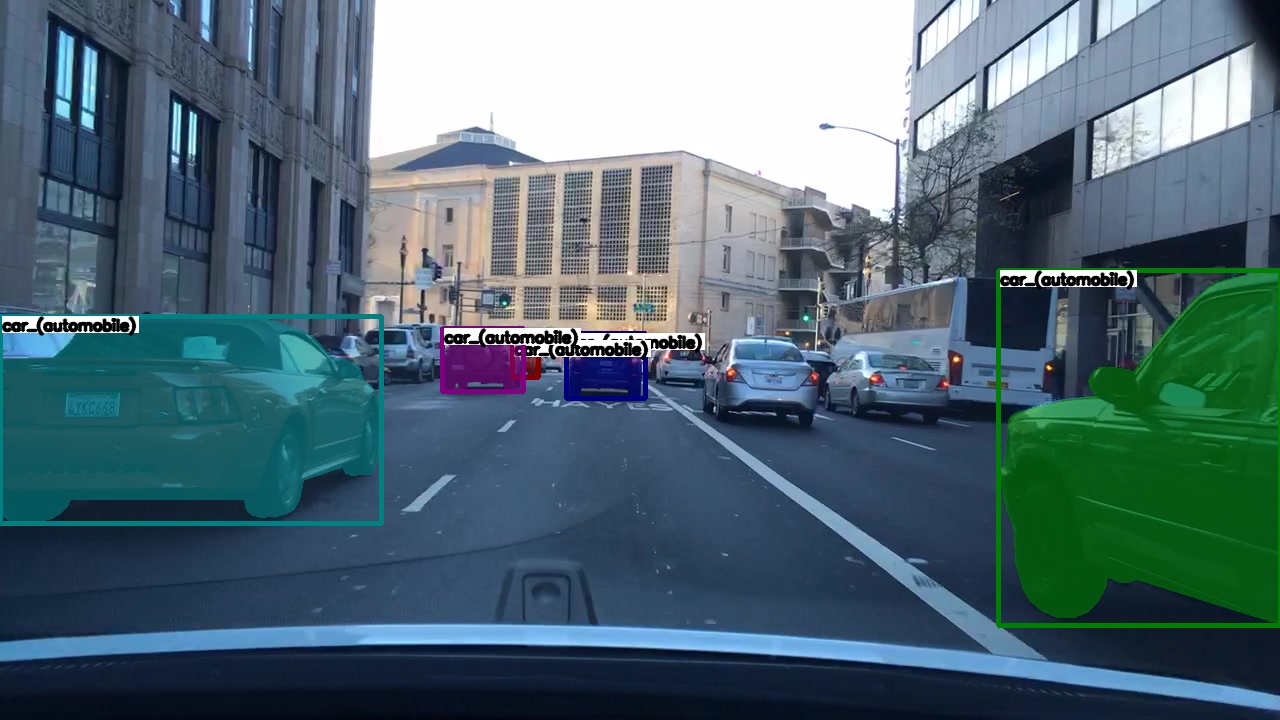}&%
       \includegraphics[width=\cellWidth]{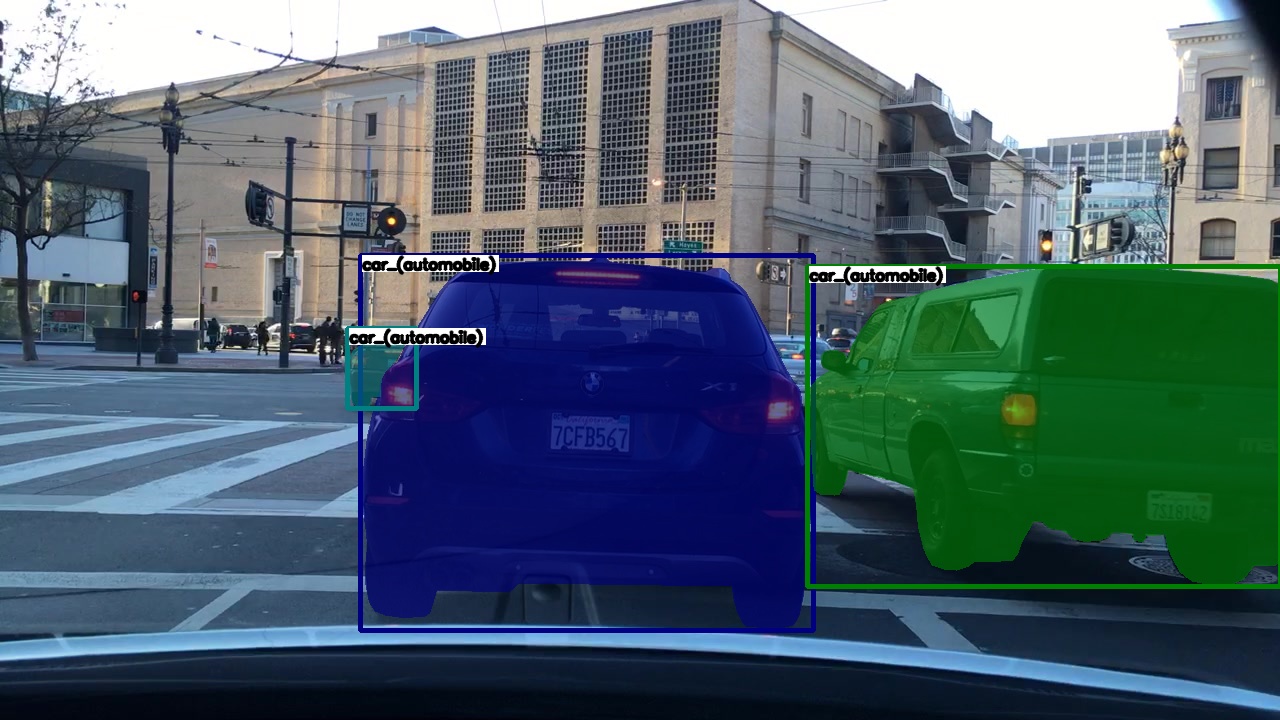} \\
  \end{tabularx}
  \caption{Assorted object annotations from \DatasetAbbrev{} showing diverse outdoor, indoor and driving scenes.}
  \label{fig:teaser}
\end{figure}

Segmenting and tracking multiple objects in video is widely researched because of applications in autonomous robots and self-driving vehicles. Over time however, this broadly defined task has splintered into multiple datasets and benchmarks, each with its own sub-community. Even though tasks such as Video Object Segmentation (VOS) and Multi-Object Tracking and Segmentation (MOTS) are closely related, there is a lack of interaction between their sub-communities.

\begin{figure*}
\centering
\includegraphics[width=5.5in]{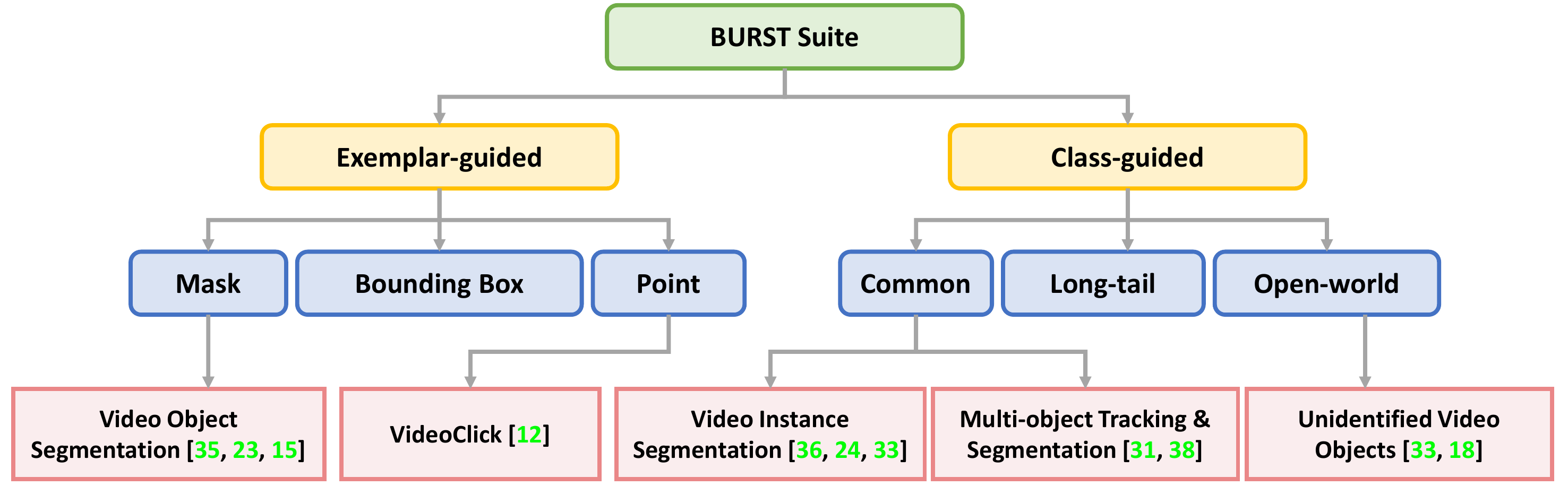}
\caption{\textbf{\DatasetAbbrev{} Task Taxonomy/Hierarchy.} Boxes in the bottom row give examples of existing benchmarks which tackle that task.}
\label{fig:task_taxonomy}
\end{figure*}

Our work aims to remedy this; we propose \DatasetAbbrev{}: a dataset containing a large, diverse set of videos with object mask annotations, and an associated benchmark with six related tasks.
\DatasetAbbrev{} is based on the existing TAO dataset by Dave~\etal~\cite{Dave20Tao} for bounding-box level multi-object tracking, but has been extensively re-annotated with pixel-precise masks.
%
%
The videos in our dataset include indoor and outdoor scenes, `in-the-wild' videos, scripted movie scenes, and street scenes captured from moving vehicles. Examples can be seen in Fig.~\ref{fig:teaser}. The six tasks in our benchmark are organized into a hierarchical taxonomy which is illustrated in Fig.~\ref{fig:task_taxonomy}. All tasks fall under the umbrella of requiring pixel-precise segmentation and tracking of potentially multiple objects in video sequences. 

%
The first level in our task hierarchy splits tasks based on the set of target objects that have to be segmented/tracked. For \emph{exemplar-guided} tasks, an explicit cue is given for each of the target objects. For \emph{class-guided} tasks, the set of target objects are all those which belong to a predefined set of object classes. The exemplar-guided stream is further divided into three tasks where, for the first video frame in which the target object appears, we are given either (i) the object mask, (ii) its bounding box or (iii) a random point inside the object. The class guided stream is also further divided into three tasks where the pre-defined class set is either (i) a small set of common object classes, (ii) a larger set of classes with several infrequently occurring classes (\ie long-tail), or (iii) an `open-world' task~\cite{liu22OWTB} where methods are trained on a small set of known classes, but during inference are expected to additionally track and segment objects belonging to a larger, previously unseen set of classes.

Fig.~\ref{fig:task_taxonomy} (bottom row) shows which existing benchmarks map onto our task taxonomy \eg Video Object Segmentation~\cite{PontTuset17Davis,Xu18YouTubeVOS} is identical to our mask exemplar-guided task, whereas Video Instance Segmentation (VIS)~\cite{Yang19YouTubeVIS,Qi21OVIS} and Multi-Object Tracking and Segmentation (MOTS)~\cite{Voigtlaender19mots} are similar to the common class-guided task. 

The hierarchy shows that these tasks are highly related to one another. As we will show in Sec.~\ref{sec:baselines}, research advances targeting one task can be utilized for other tasks.
%
%
For example, state-of-the-art methods~\cite{Cheng21stcn,Yang21aot,oh19STMVOS,seong21HMNVOS} for exemplar-guided tasks work by `propagating' object masks from one video frame to another. Here we note that improvements in mask-propagation (\ie temporal association) can benefit class-guided methods.
Conversely, several existing methods~\cite{bergmann19Traktor,Tang17TrackByDetection,Luiten20Unovost,pavel21TbD} for class-guided tasks employ image-level object detectors to first segment objects individual video frames. Here we note that improvements to image-level object detectors could also benefit methods tackling the exemplar-guided task.
%
With \DatasetAbbrev{}, we aim to bring together methods for these tasks under a single umbrella benchmark to encourage more knowledge exchange.
%
%
To further facilitate unification and interaction, we use the same set of metrics based on Higher-Order Tracking Accuracy (HOTA)~\cite{luiten2020IJCV} for all tasks. This enables direct, quantitative comparison between different methods targeting different tasks. To demonstrate the usefulness of this feature, we setup several effective baselines for our proposed tasks, some of which are constructed by bootstrapping standard approaches for other tasks. The comparability of the resulting scores offers interesting insight into how well methods generalize across tasks.


To summarize, we propose \DatasetAbbrev{}: a large, diverse and challenging dataset with mask-level object annotations, and an associated benchmark with 6 tasks related to segmenting and tracking multiple objects in video. Methods can be evaluated for one or more tasks using the same underlying data and comparable metrics. This aims to encourage greater cohesion and knowledge exchange between researchers working on these tasks, 
and accelerate development of generalized methods that can tackle multiple tasks. 
%
%

\section{The \DatasetAbbrev{} Benchmark}
\label{sec:dataset}

Existing object tracking and segmentation datasets are typically geared towards certain types of video scenes \eg in-the-wild internet videos~\cite{PontTuset17Davis,Yang19YouTubeVIS,Xu18YouTubeVOS,Qi21OVIS}, outdoor street scenes captured from a driving vehicle~\cite{Geiger12KITTI,Yu20bdd100k,Chang19argoverse,homayounfar2021videoclick}. 
The videos in \DatasetAbbrev{}, on the other hand, cover multiple types of scenes
and encompasses a large set of \numclasses{} object classes. 
%
We use the videos from TAO~\cite{Dave20Tao}, which is in-turn composed from videos belonging to 7 different datasets: ArgoVerse~\cite{Chang19argoverse} and BDD~\cite{Yu20bdd100k}, which contain outdoor driving scenes captured from moving vehicles, LaSOT~\cite{Fan19lasot} and YFCC100M~\cite{Thomee16yfcc100m}, which contain assorted, in-the-wild videos from the internet, and AVA~\cite{Gu18ava}, Charades~\cite{Sigurdsson16charades} and HACS~\cite{Zhao19hacs}, which contain videos with human-human and human-object interactions, but with some subtle differences: Charades contains mostly indoor scenes with slow object motion, AVA contains snippets from scripted movies, and HACS contains in-the-wild internet videos. 

\DatasetAbbrev{} contains 2,914 videos with a lower frame dimensions of at least 480px. The videos are $\sim$30s in length, and
the training, validation and test set contain 500, 993, and 1421 videos, respectively. The training set is annotated at 6fps whereas both validation and test sets are annotated at 1fps. 
Table~\ref{tab:basic_dataset_stats} summarizes statistics for \DatasetAbbrev{}.


\begin{table}[h]
    \centering
    
    \caption{Statistics for \DatasetAbbrev{} train, validation and test sets.}
    \vspace{5pt}

    \resizebox{\linewidth}{!}{
    \begin{tabular}{lcccc}
    \toprule 
    & Train & Validation & Test & Total\tabularnewline
    \midrule 
    Annotation fps & 6 & 1 & 1 & -\tabularnewline
    Videos & 500 & 993 & 1421 & 2,914 \tabularnewline
    Total video length (hrs) & 4.94 & 9.84 & 14.12 & 28.9 \tabularnewline
    Object tracks & 2,645 & 5,481 & 7,963 & 16,089 \tabularnewline
    Annotated frames & 107,144 & 36,375 & 52,194 & 195,713 \tabularnewline
    Object masks & 318,200$^1$ & 114,825 & 167,132 & 600,157\tabularnewline
    \bottomrule
    \multicolumn{5}{l}{$^1$ includes 212,477 automatically generated and consistency-verified masks}
    \end{tabular}
    }

    
    \label{tab:basic_dataset_stats}
\end{table}

\PAR{Federated Annotations.} 
Similar to TAO~\cite{Dave20Tao}, the annotations in \DatasetAbbrev{} are \textit{federated}, \ie not all objects belonging to the predefined set of object classes are annotated in every video. This is
similar to the philosophy behind the image-level LVIS~\cite{Gupta19LVIS} and OpenImages~\cite{Luznetsova20OpenImages} datasets.
Every video in \DatasetAbbrev{} contains the following, in addition to the non-exhaustive annotations: (1) a list of object classes which are non-exhaustively annotated, and (2) a list of object classes which are not present in the video.
This information enables us to derive three sets of videos for every object class: where it is present, absent and non-exhaustively present.
This in-turn is used to penalize false positives and negatives for each object class during evaluation.
We refer readers to the TAO dataset paper~\cite{Dave20Tao} for more details.

\section{Comparison to Related Datasets}
\label{sec:comparison}

There exist several datasets of various sizes which tackle one or more tasks evaluated by our benchmark. Tables~\ref{tab:dataset_comparison_by_difficulty} and~\ref{tab:dataset_comparison_by_task} compare \DatasetAbbrev{} to these existing datasets in a number of ways, as detailed in the following sub-sections. 


\subsection{Comparison by Tasks}

Table~\ref{tab:dataset_comparison_by_task} shows which tasks each dataset/benchmark evaluates in terms of our task taxonomy (Fig.~\ref{fig:task_taxonomy}).
We note that existing benchmarks typically address one or at most two tasks. 
The `{\color{orange}/}' entries mean that the dataset does not evaluate the given task, but that it is possible to do so \eg a class-guided benchmark can also be formulated as an exemplar-guided benchmark by assuming that the first-frame object masks are known during inference.

For exemplar-guided tasks, the two most common benchmarks are DAVIS~\cite{PontTuset17Davis} and YouTube-VOS~\cite{Xu18YouTubeVOS}. Both contain diverse, in-the-wild, videos from the internet which are $\sim$5-10s in length. VOT~\cite{Kristan18VOT18} contains longer videos, but is a single-object tracking dataset.

Looking at the class-guided stream, most datasets can be assigned to one of two distinct groups. On one hand, benchmarks such as BDD~\cite{Yu20bdd100k}, KITTI~\cite{Geiger12KITTI} and MOTS-Challenge~\cite{Voigtlaender19mots} are inspired from classical Multi-Object Tracking (MOT). They target autonomous driving problems, 
and contain lengthy videos of street scenes captured from a driving vehicle or a walking pedestrian. We see that KITTI and MOTS-Challenge evaluate methods using `sMOTSA', which is an extension of the popular MOTA (Multi-Object Tracking Accuracy) measure~\cite{Stiefelhagen07MOTA} for when segmentation masks are used instead of bounding-boxes.  On the other hand, datasets like YouTube-VIS~\cite{Yang19YouTubeVIS} and UVO~\cite{Wang21UVO} appear more related to Video Object Segmentation (VOS), and typically contain diverse, but shorter videos from the internet. OVIS~\cite{Qi21OVIS} can be seen as an extension of YouTube-VIS with longer videos and more object occlusion. 
UVO stands out from the others in that it could be used for the open-world task since it contains mask annotations for a anything which humans would consider to be `objects' as opposed to being restricted to a small class of objects. Datasets in this category use mean Average Precision (mAP) as an evaluation measure. These benchmarks can thus be seen as video extensions of image-level instance segmentation benchmarks such as COCO~\cite{Lin14coco}, LVIS~\cite{Gupta19LVIS} and OpenImages~\cite{Luznetsova20OpenImages} where mAP is the metric of choice. 

In contrast to all of the above, \DatasetAbbrev{} contains the required annotations to evaluate all six tasks. In particular, the long-tail open-world tasks are enabled by the fact that our object class set is sufficiently large.

\begin{table*}
\centering{}
\setlength{\tabcolsep}{1.3pt}
\renewcommand{\arraystretch}{1.0}
\scriptsize

\caption{\textbf{Dataset Comparison By Size and Difficulty.} Comparison of datasets according to various measures of `difficulty'. Statistics for validation/test may not be exact if data is not publicly available.
\label{tab:dataset_comparison_by_difficulty}
}   
\vspace{5pt}

\begin{tabular}{cccccccccccccccc}
\toprule 
 &  & \multicolumn{4}{c}{\textbf{Difficulty}} &  & \multicolumn{4}{c}{\textbf{Train Size}} &  & \multicolumn{4}{c}{\textbf{Validation / Test Size}}\tabularnewline
\cmidrule{1-1} \cmidrule{3-6} \cmidrule{8-11} \cmidrule{13-16} 
Dataset &  & Setting & Length (hrs) & Masks / Frame & \# Classes &  & Ann Masks & Ann Tracks & Ann Frames & Ann Vids &  & Ann Masks & Ann Tracks & Ann Frames & Ann Vids\tabularnewline
\cmidrule{1-1} \cmidrule{3-6} \cmidrule{8-11} \cmidrule{13-16} 
VOT~\cite{Kristan18VOT18} &  & Single Object & 10.7 & 1 & - &  & 0 & 0 & 0 & 0 &  & 19,903 & 62 & 19,903 & 62\tabularnewline
DAVIS`17~\cite{PontTuset17Davis} &  & Internet Videos & 2.9 & 2.6 & 78 &  & 10,238 & 144 & 4,219 & 60 &  & 16,841 & 242 & 6,240 & 90 \tabularnewline
YT-VOS~\cite{Xu18YouTubeVOS} &  & Internet Videos & 4.5 & 1.63 & 94 &  & 12,918 & 6,459 & 94,440 & 3,471 &  & 4,310 & 2,155 &  28,825 & 1,048 \tabularnewline
BDD~\cite{Yu20bdd100k} &  & Driving & 40 & 11.4 & 7 &  & 347,442 & 17,838 & 30,745 & 154 &  & 77,389 & 4,873 & 6,475 & 32\tabularnewline
KITTI-MOTS~\cite{Voigtlaender19mots} &  & Driving & 39.0 & 5.2 & 2 &  & 38,197 & 748 & 8,008 & 21 &  & 61,906 & 961 & 11,095 & 28\tabularnewline
MOTS-Chal.~\cite{Voigtlaender19mots} &  & Surveillance & 34.4 & 10.0 & 1 &  & 26,894 & 228 & 2,862 & 4 &  & 32,269  & 328 & 3044 & 4\tabularnewline
YT-VIS~\cite{Yang19YouTubeVIS} &  & Internet Videos & 4.5 & 1.69 & 40 &  & 103,424 & 3,774 & 61,845 & 2,238  &  & 29,431 & 1,092 & 17,415  & 645\tabularnewline
UVO~\cite{Wang21UVO} &  & Human Actions & 3 & 12.3 & - &  & 416,001 & 76,627 & 39,174 & 5,641 &  & 177,153 & 28,271 & 18,966 & 5,587\tabularnewline
\addlinespace[1pt]\hline\addlinespace[2pt]
\DatasetAbbrev{} &  & General / Diverse & 28.9 & 3.1 & \numclasses{} &  & 318,200 & 2,645 & 107,144 & 500 &  & 281,957 & 13,444 & 88,569 & 2,414\tabularnewline
\bottomrule
\end{tabular}

\end{table*}

\begin{table}
\centering{}
\setlength{\tabcolsep}{1.3pt}
\renewcommand{\arraystretch}{1.0}

\caption{\textbf{Dataset Comparison by Task.} `{\color{orange}/}' means that the dataset contains annotations to setup the given task, but this is not done officially as part of that benchmark.
\label{tab:dataset_comparison_by_task}
}
\vspace{5pt}

\resizebox{\linewidth}{!}{
\begin{tabular}{lccc@{\hskip 2em}ccc}
\toprule 
 &  \multicolumn{3}{c}{Exemplar-guided} & \multicolumn{3}{c}{Class-guided} \\
 &  Mask & Box & Point & Common & Long-Tail & Open-World\tabularnewline
\midrule 
VOT~\cite{Kristan18VOT18}   & {\color{customgreen}\checkmark} & {\color{customgreen}\checkmark} & {\color{orange}/} & {\color{Maroon}\xmark} & {\color{Maroon}\xmark} & {\color{Maroon}\xmark}\tabularnewline
DAVIS~\cite{PontTuset17Davis}   & {\color{customgreen}\checkmark} & {\color{orange}/} & {\color{orange}/} & {\color{Maroon}\xmark} & {\color{Maroon}\xmark} & {\color{Maroon}\xmark}\tabularnewline
YouTube-VOS~\cite{Xu18YouTubeVOS}   & {\color{customgreen}\checkmark} & {\color{orange}/} & {\color{orange}/} & {\color{Maroon}\xmark} & {\color{Maroon}\xmark} & {\color{Maroon}\xmark}\tabularnewline
BDD~\cite{Yu20bdd100k}   & {\color{orange}/} & {\color{orange}/} & {\color{orange}/} & {\color{customgreen}\checkmark} & {\color{Maroon}\xmark} & {\color{Maroon}\xmark}\tabularnewline
KITTI-MOTS~\cite{Voigtlaender19mots}   & {\color{orange}/} & {\color{orange}/} & {\color{orange}/} & {\color{customgreen}\checkmark} & {\color{Maroon}\xmark} & {\color{Maroon}\xmark}\tabularnewline
MOTS-Chal.~\cite{Voigtlaender19mots}   & {\color{orange}/} & {\color{orange}/} & {\color{orange}/} & {\color{customgreen}\checkmark} & {\color{Maroon}\xmark} & {\color{Maroon}\xmark}\tabularnewline
YouTube-VIS~\cite{Yang19YouTubeVIS}   & {\color{orange}/} & {\color{orange}/} & {\color{orange}/} & {\color{customgreen}\checkmark} & {\color{Maroon}\xmark} & {\color{Maroon}\xmark}\tabularnewline
OVIS~\cite{Qi21OVIS}   & {\color{orange}/} & {\color{orange}/} & {\color{orange}/} & {\color{customgreen}\checkmark} & {\color{Maroon}\xmark} & {\color{Maroon}\xmark}\tabularnewline
UVO~\cite{Wang21UVO}   & {\color{orange}/} & {\color{orange}/} & {\color{orange}/} & {\color{customgreen}\checkmark} & {\color{Maroon}\xmark} & {\color{customgreen}\checkmark} \tabularnewline
\addlinespace[1pt]\hline\addlinespace[2pt]
\DatasetAbbrev{}   & {\color{customgreen}\checkmark} & {\color{customgreen}\checkmark} & {\color{customgreen}\checkmark} & {\color{customgreen}\checkmark} & {\color{customgreen}\checkmark} & {\color{customgreen}\checkmark}\tabularnewline
\bottomrule
\end{tabular} 
}

\end{table}

\subsection{Comparison by Difficulty}

Table~\ref{tab:dataset_comparison_by_difficulty} lists several datasets along with various parameters which subjectively determine their `difficulty'. In terms of video length, the average sequence in \DatasetAbbrev{} lasts 36.8s, which is longer than the other datasets. Lengthy videos are challenging because of more object instances, longer occlusions, and are also more memory demanding because of the higher frame count. 
In terms of number of annotations, \DatasetAbbrev{} contains a total of $\sim$600k object masks across $\sim$200k video frames, which is larger than most other datasets except BDD and UVO.
%
With regard to object classes, \DatasetAbbrev{} contains objects belonging to \numclasses{} possible classes, which is significantly higher than the class set for other benchmarks. To the best of our knowledge, we are the first to provide pixel-precise object annotations for such a large set of object classes. 
Besides increasing object diversity, this feature also enables us to evaluate methods for the long-tail class-guided task. As mentioned in Sec.~\ref{sec:dataset} however, our annotations are federated, \ie not exhaustive.
Finally, we note that that \DatasetAbbrev{} can better evaluate the generalization capability of methods since it contains more scene diversity compared to existing datasets, many of which focus on specific settings \eg driving scenes.



\section{Dataset Creation}
\label{sec:creation}



We build upon the TAO dataset~\cite{Dave20Tao} which contains object bounding-box annotations at 1fps. 
We professionally re-annotated this to obtain pixel-precise masks for all 342,052 object bounding-boxes\footnote{We re-used the small set of 27,500 masks published by \cite{voigtlaender21WACV}.}.
%
%
We then set out to increase the temporal density of annotations in the training set from 1fps to 6fps.
Visualizing a sequence of annotations at 1fps shows large movements and appearance changes between successive frames. It is challenging to train tracking-related methods on this data since they are designed to learn video motion cues from smooth video frame progression. However, annotating object masks at the full video frame-rate (24-30fps) would be in-feasibly expensive and highly redundant since there is usually little scene change between successive frames. Annotating at 6fps is thus a compromise (also used by other datasets~\cite{PontTuset17Davis,Xu18YouTubeVOS,Yang19YouTubeVIS}) since it reduces annotation cost while still ensuring smooth scene progression.
For \DatasetAbbrev{} however, even 6fps annotations would require 255,654 additional mask annotations for the training set. To reduce cost and human effort, we instead developed a semi-automatic procedure to do this, as explained below:

\begin{figure*}
\centering
\definecolor{darkbrown}{rgb}{0.5, 0.20, 0.00} 

\includegraphics[width=0.85\linewidth]{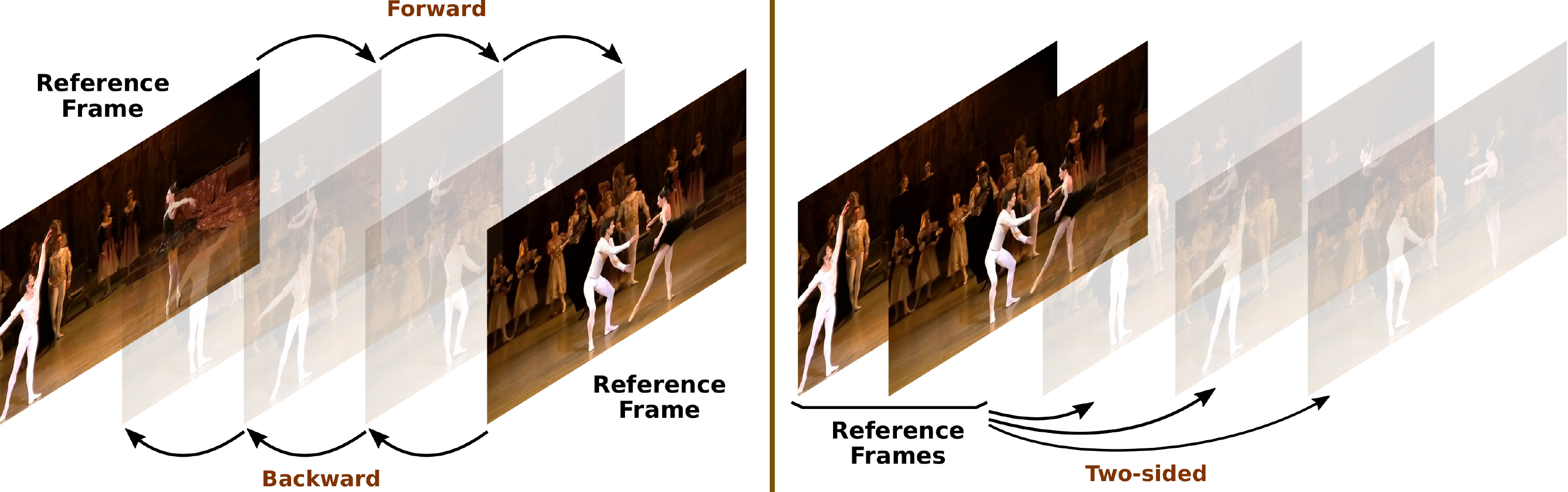}
\caption{Illustration of mask propagation techniques for densifying the training set. STCN~\cite{Cheng21stcn} and AOT-L~\cite{Yang21aot} are both executed in the \textcolor{darkbrown}{forward} and \textcolor{darkbrown}{backward} settings. The \textcolor{darkbrown}{two-sided} setting is only executed for STCN. 
}
\label{fig:ensemble_methods_for_densification}
\end{figure*}

\PAR{1. Automatic Mask Propagation}
Interestingly, the task of temporally densifying annotations is practically identical to the mask exemplar guided task referred to in Fig.~\ref{fig:task_taxonomy}: given an object mask in a certain frame, we need masks for the same object in other video frames. In this case, the video length over which this mask propagation step has to be performed is quite short -- at most 1s since we already have human-labeled 1fps annotations. We found that two recent, state-of-the-art methods for `Video Object Segmentation', namely STCN~\cite{Cheng21stcn} and AOT-L~\cite{Yang21aot}, perform this task quite well with off-the-shelf trained weights. To further improve mask quality, we obtain two different sets of results from each of these methods by running them in two different ways. 
As shown in Fig.~\ref{fig:ensemble_methods_for_densification} (left), given a pair of annotated frames with a number of non-annotated frames in between, we can either run the method with the first frame as the reference and propagate \emph{forward} sequentially, or, starting from the last frame as reference, propagate \emph{backward} sequentially. Doing so for both methods results in a total of 4 different sets of propagated masks. Additionally, we use STCN to obtain a fifth, tie-breaking \emph{two-sided} result by using both annotated frames as the reference frames (Fig.~\ref{fig:ensemble_methods_for_densification}, right), and propagating the masks directly to each of the other non-annotated frames (\ie the frame history update mechanism in STCN is disabled). Thus, we have a total of 5 masks for each object. We subsequently perform a per-pixel majority vote to obtain a final \textit{consensus mask}.

\PAR{2. Mask Quality Assessment}
Although most annotations produced by step 1 are high quality, there are several failure cases \eg poor lighting, occluded scenes, erratic camera motion. To identify them, one could manually inspect each object mask and decide if it is of ground-truth quality. Even though doing this is still significantly less costly than fully annotating the object masks, we nonetheless developed a more efficient yet effective procedure for evaluating mask quality: we compute the IoU of each of the five masks generated in step 1 with the \emph{consensus mask}. The five resulting IoUs are then averaged to obtain a final metric in $[0,1]$ which is treated as a quality score $Q$ for the \textit{consensus mask}.

\begin{figure}
\centering
\includegraphics[width=\linewidth]{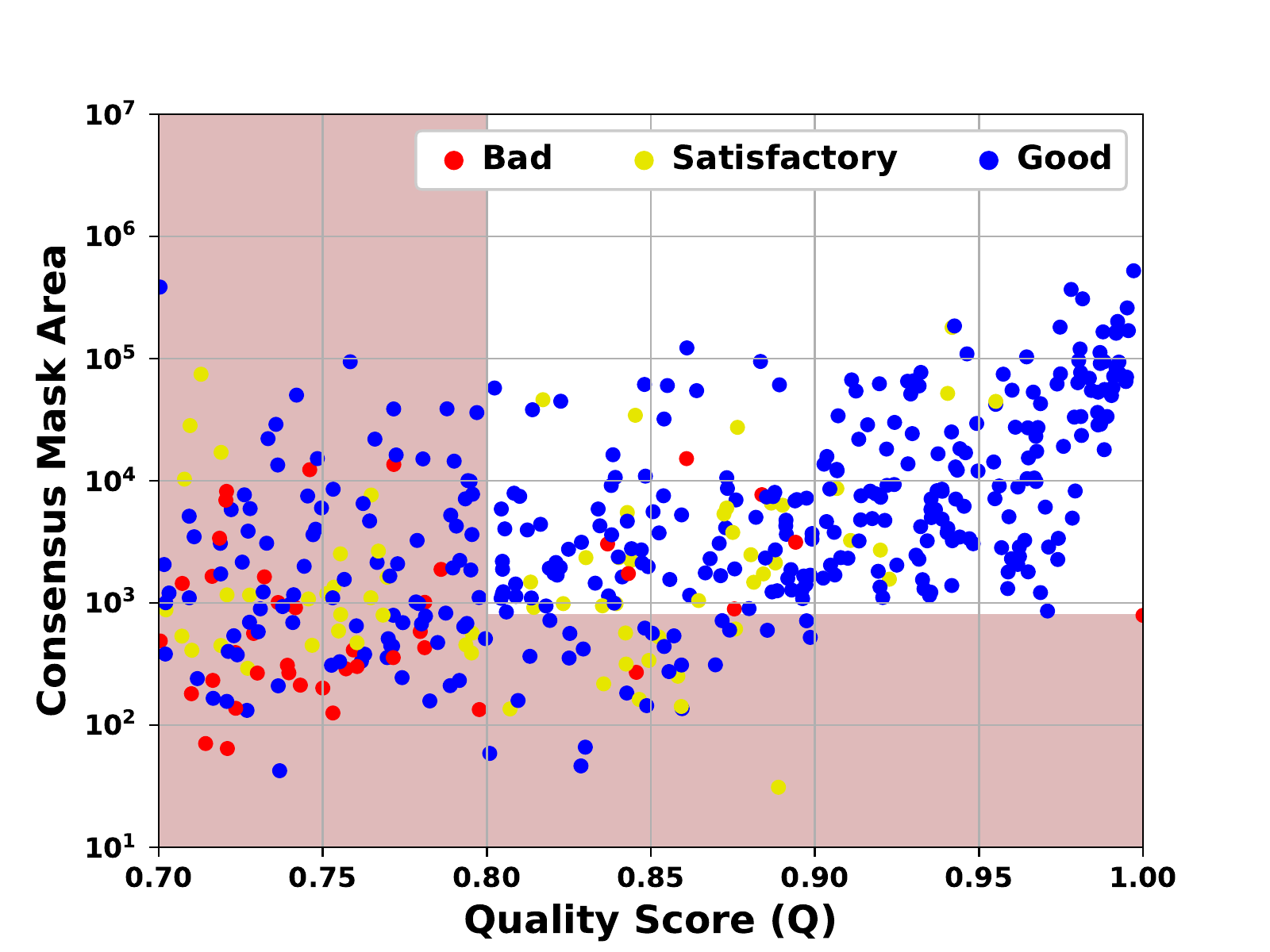}
\caption{Human assessment of mask quality plotted against our automated quality score $Q$ and pixel area of the consensus masks. All masks falling in the red region were manually re-annotated.
}
\label{fig:auto_annotated_mask_quality_assessment}
\end{figure}

\PAR{3. Manual Re-annotation of Low Quality Masks.} To decide which masks are of sub-standard quality, we consider two key measures: the score $Q$ from step 2, and the pixel mask area of the consensus mask. We asked two professional annotators to manually assess the quality of a set of 250 object masks. These were sampled such that they are uniformly distributed with respect to their $Q$ scores and pixel areas. The annotators were asked to assign one of three ratings to each object mask: (1) `good': mask quality is as good as human annotated ground-truth, (2) `satisfactory': there are visible errors \eg object contours are imperfect, minor instances of mask fragmentation, but overall still acceptable and (3) `bad': there are unacceptable errors \eg object ID switches, gross under/over-segmentation. Fig.~\ref{fig:auto_annotated_mask_quality_assessment} illustrates the results of this survey: each object mask is shown as a point whose color reflects the human-assigned rating. The points are plotted w.r.t their $Q$ score and mask pixel areas. We observe a strong correlation between both of these measures and the human-perceived mask quality, since most `bad' masks are in the lower-left corner of the plot, and vice versa. Based on this plot, we decided to manually re-annotate all object masks whose $Q$ scores were below 0.8, or whose mask areas were smaller than 750 pixels. This region is highlighted in red in Fig.~\ref{fig:auto_annotated_mask_quality_assessment}.

By using this workflow to densify the training set annotations from 1fps to 6fps, we only require 43,177 out of 255,654 object masks (16.9\%) to be manually annotated since the remaining automatically generated masks passed the quality threshold to be considered ground-truth.

\section{\DatasetAbbrev{} Task Taxonomy}
\label{sec:tasks}

As explained in Sec.~\ref{sec:intro}, 
several existing benchmarks involve closely related tasks pertaining to segmenting and tracking multiple objects in video, but
there is limited cross-interaction between their respective research sub-communities. With \DatasetAbbrev{}, we aim to unify these disparate benchmarks under a single umbrella with shared data and consistent evaluation metrics. The six tasks constituting \DatasetAbbrev{} are illustrated in Fig.~\ref{fig:task_taxonomy} and explained below.


\subsection{Exemplar-guided}

This set of tasks requires tracking and segmenting multiple target objects in a video given some ground-truth cue for each of these objects in the first video frame in which they appear. Note that this may not necessarily be the first frame of the video. The three tasks in this stream are based on the type of the given cue:

\PAR{1) Mask.}
The method is given the segmentation mask for each of the target objects in the first frame. 

\PAR{2) Box.}
The bounding box coordinates of the target objects in the first frame are given. Note that the predicted output should still be pixel-precise segmentation masks.

\PAR{3) Point.}
This is the most challenging of the three where the method is only given one pixel coordinate which lies inside the target object mask. Again, the predicted output should still contain segmentation masks.

\subsection{Class-guided}

For this set of tasks, methods are required to track, segment and assign a class label to all objects in a video which belong a pre-defined set of object classes. 
The three tasks in this stream are:

\PAR{4) Common.}
Here, the target class set includes 78 classes from the popular COCO dataset~\cite{Lin14coco} spanning diverse object categories \eg animals, persons, vehicles, furniture, food items. 

\PAR{5) Long-tailed.}
This task involves a large set of \numclasses{} object classes from the LVIS dataset~\cite{Gupta19LVIS}. It is challenging because several classes contain very few training samples.

\PAR{6) Open-world.}
The idea behind open-world instance segmentation~\cite{liu22OWTB} is that methods are trained on a certain, `known' set of object classes, but during inference, they are expected to additionally segment objects belonging to an `unknown' class set. Methods need not assign class labels to the predicted instances, and the evaluation does not penalize false positives. For our open-world task, the 78 `common' classes are the `known' set, and the `unknown' set includes everything that is in the 482 class `long-tail' set, but not in the `common' set.








\section{Unified Evaluation Metrics}
\label{sec:metrics}


We evaluate all tasks using Higher Order Tracking Accuracy (\hota{})~\cite{luiten2020IJCV} because it strikes a good balance between measuring frame-level detection and temporal association accuracy. For the open-world task, a slightly modified, recall-based variant of \hota{} called Open World Tracking Accuracy (OWTA) is used.

\PAR{HOTA.}
To calculate \hota{}~\cite{luiten2020IJCV}, the predicted detections (per-frame) are first matched to the ground truth detections based on the IoU between their masks. 
Using this mapping, the Detection Accuracy (DetA) and the Association Accuracy (AssA) can be calculated and combined by taking their geometric mean to obtain the HOTA score, \ie
\begin{equation}
    \mathrm{HOTA} = \sqrt{\mathrm{DetA} \cdot \mathrm{AssA}}.
\end{equation}

Multiple IoU thresholds are used to compute the prediction $\xleftrightarrow{}$ ground-truth matching; the final \hota{} (and $\mathrm{DetA}$, $\mathrm{AssA}$) is calculated by averaging over the thresholds.

\PAR{Detection Accuracy (DetA).}
Using the mapping between predicted and ground truth detections, these detections can be partitioned into a set of True Positive (TP), False Positive (FP), and False Negative (FN) detections. 
The DetA, which solely measures the quality of the detections while disregarding track associations, can then be obtained by
\begin{equation}
    \mathrm{DetA} = \frac{|\mathrm{TP}|}{|\mathrm{TP}|+|\mathrm{FN}|+|\mathrm{FP}|}.
\end{equation}

\PAR{Association Accuracy (AssA).}
To calculate the AssA, an association score $\mathcal{A}(c)$ is calculated for each true positive detection $c$. The final AssA score is obtained by averaging over the set of true positive detections $\mathrm{TP}$:
\begin{equation}
    \mathrm{AssA} = \frac{1}{|\mathrm{TP}|} \sum_{c \in \{\mathrm{TP}\}} \mathcal{A}(c).
\end{equation}
The association score $\mathcal{A}(c)$ for true positive detection $c$ is calculated as
\begin{equation}
    \mathcal{A}(c) = \frac{|\mathrm{TPA}(c)|}{|\mathrm{TPA}(c)| + |\mathrm{FNA}(c)| + |\mathrm{FPA}(c)|},
\end{equation}
where True Positive Associations (TPAs), False Positive Assocations (FPAs), and False Negative Associations (FNAs) \cite{luiten2020IJCV} are computed by comparing the whole predicted track which goes through detection $c$ with the whole ground truth track which goes through detection $c$. We refer the reader to Luiten~\etal~\cite{luiten2020IJCV} for detailed explanations. 

\PAR{Object Classes.}
To handle multiple object classes, \hota{} can be calculated separately for each class, followed by an averaging step to yield a final metric. To facilitate easier performance analysis over different object classes, we average the per-class \hota{} scores over three different sets of object classes: (1) `common' set, which contains 78 object classes from COCO~\cite{Lin14coco}, (2) `uncommon' set, which contains 404 infrequently occurring object classes from LVIS~\cite{Gupta19LVIS}, and (3) `all' set, which is the union of both ($\text{78}+\text{404}=\text{\numclasses{}}$ classes).
We denote these three metrics with \hotacommon{}, \hotauncommon{} and \hotaall{}, respectively.
%

\PAR{HOTA for Exemplar-guided Tasks.}
The evaluation for the exemplar-guided task is identical to the class guided task, and the scores can be directly compared. However, it should be noted that the exemplar-guided methods inherently receive extra ground-truth information: a mask/box/point, and the class label for each target object.

\subsection{Open-world Evaluation}
For the open-world task, methods are expected to segment and track objects of previously unseen classes. Since it is infeasible to label every single object (even the definition of `object' is ambiguous), we have to assume that the prediction may contain valid objects which are not covered by the ground truth.
This entails that false positive detections should not be penalized, and hence, for the open-world task, we replace HOTA with Open-World Tracking Accuracy (OWTA) \cite{liu22OWTB}, which is calculated as:
\begin{equation}
    \mathrm{OWTA} = \sqrt{\mathrm{DetRe} \cdot \mathrm{AssA}},
\end{equation}

where the Detection Recall ($\mathrm{DetRe}$) is given by

\begin{equation}
    \mathrm{DetRe} = \frac{|\mathrm{TP}|}{|\mathrm{TP}|+|\mathrm{FN}|},
\end{equation}

Note that DetRe is similar to DetA, but it disregards false positives (FP). To prevent methods from obtaining a high score by simply predicting an extremely large number of detections, we mandate that object mask predictions for the open-world task cannot overlap with each other.

\section{Baselines}
\label{sec:baselines}

\begin{table*}
\centering{}
\setlength{\tabcolsep}{4.0pt}
\newcommand\RotText[1]{\rotatebox{90}{\parbox{2cm}{\centering#1}}}
\footnotesize

\caption{
Baseline results for all tasks using various methods.
Evaluation metrics are reported separately for `common', `uncommon' and `all' classes. 
Object detector training data: \textbf{*}: COCO, \textbf{$\dagger$}: LVIS. 
}
\vspace{4pt}

\newcommand{\tc}[1]{\multicolumn{3}{c}{\textbf{#1}}}
\newcommand{\ha}{$\mathrm{HOTA}_\text{all}$}
\newcommand{\hc}{$\mathrm{HOTA}_\text{com}$}
\newcommand{\hu}{$\mathrm{HOTA}_\text{unc}$}
\newcommand{\oa}{$\mathrm{OWTA}_\text{all}$}
\newcommand{\oc}{$\mathrm{OWTA}_\text{com}$}
\newcommand{\ou}{$\mathrm{OWTA}_\text{unc}$}
\newcommand{\task}[2]{\multirow{#2}{7em}{\centering #1}}
\newcommand{\stask}[2]{\multirow{#2}{3em}{\RotText{#1}}}

\begin{tabular}{cclcccccccc} 
\toprule 
                                &                       &                           &   & \tc{Validation}                  &        & \tc{Test}                 \\
                                &                       & Baseline Method           &   & \ha   & \hc   & \hu       &   & \ha       & \hc       & \hu      \\
\cmidrule{2-3} \cmidrule{5-7} \cmidrule{9-11}                                                                                                  
\stask{Exemplar\\ Guided}{7}    & \task{Mask}{2}        & STCN~\cite{Cheng21stcn}   &   & 49.8  & 52.2  & 49.2      &   &   52.4    &   51.1     &   52.7     \\ 
                                &       & Box Tracker{*}~\cite{luiten2020trackeval} &   & 18.0  & 35.8  & 13.6      &   &   14.1    &   28.0     &   11.4     \\ 
\cmidrule{2-3} \cmidrule{5-7} \cmidrule{9-11}                                                                                                                         
                                & \task{Box}{3}         & STCN (PointRend)          &   & 45.2  & 48.9  & 44.3      &   &   46.0    &   48.9     &   45.4    \\ 
                                &                       & STCN (Matched Det{*})     &   & 24.5  & 47.6  & 18.7      &   &   25.0    &   41.9     &   21.7    \\ 
                                &                       & Box Tracker{*}            &   & 13.7  & 34.2  & 8.6       &   &   13.6    &   27.7     &   10.8    \\ 
\cmidrule{2-3} \cmidrule{5-7} \cmidrule{9-11}                                                                                                                               
                                & \task{Point}{2}       & STCN (Matched Det{*})     &   & 24.4  & 44.0  & 19.5      &   &   24.9    &   39.5     &   22.0     \\ 
                                &                       & Box Tracker{*}            &   & 12.7  & 31.7  & 7.9       &   &   10.1    &   24.4     &   7.3     \\ 
\midrule                                                                                                                                                    
\stask{Class\\ Guided}{9}       & \task{Common}{2}     & STCN Tracker{*}            &   & -     & 51.2  & -         &   &     -     &   34.6     &   -     \\  
                                &                      & Box Tracker{*}             &   & -     & 45.5  & -         &   &     -     &   34.3     &   -         \\ 
\cmidrule{2-3} \cmidrule{5-7} \cmidrule{9-11}                                                                                                                  
                                & \task{Long-tail}{2}  & STCN Tracker$^{\dagger}$   &   & 5.5   & 17.5  & 2.5       &   &    4.5    &   17.1     &  2.0      \\ 
                                &                      & Box Tracker$^{\dagger}$    &   & 8.2   & 27.0  & 3.6       &   &    5.7    &   20.1     &  2.9    \\  
\cmidrule{2-3} \cmidrule{5-7} \cmidrule{9-11}                                                                                                          
                                & \task{Open-world}{4} &                            &   & \oa   & \oc   & \ou       &   & \oa       & \oc   & \ou      \\
\cmidrule{5-7} \cmidrule{9-11}  
                                &                      & STCN Tracker               &   & 64.6  & 71.0  & 25.0      &   &    57.5   &  62.9      &  23.9    \\
                                &                      & Box Tracker                &   & 60.9  & 66.9  & 24.0      &   &    55.9   &  61.0      &  24.6     \\
                                &                      & OWTB~\cite{liu22OWTB}      &   & 55.8  & 59.8  & 38.8      &   & 56.0      &  59.9      &  38.3        \\
\bottomrule 
\end{tabular}

\label{tab:baselines}
\end{table*}

For each of the six tasks (Sec.~\ref{sec:tasks}), we implement baselines which utilize existing works with off-the-shelf trained models. Besides serving as a quantitative anchor for comparison of future works, we show how the same approaches can be utilized across multiple tasks, and how performances across tasks can be directly compared and analyzed. 

In general, we heavily utilize image-level object detectors in the context of `tracking-by-detection', where the tracking task is conceptually divided into two steps: a `detection' step in which objects are segmented in individual frames, followed by a `tracking' step in which the per-frame detections are associated over time. Although recent state-of-the-art methods~\cite{Athar20Stemseg,mahadevan2020bmvc,Wang21VisTr,Cheng21mask2former} diverge from this paradigm by jointly segmenting and tracking objects in video clips, we find that it remains a versatile approach for tackling the tasks in \DatasetAbbrev{}. We construct functional baselines for each task using some variation of tracking-by-detection. The following sub-sections detail each of the baselines and the results are presented in Table~\ref{tab:baselines}.

\subsection{Exemplar-guided}

We show two baselines for each task in this stream: (1) applying STCN~\cite{Cheng21stcn}, which is a recent `Video Object Segmentation' method for propagating object masks through video, and (2) a simple box tracker which builds object tracks by starting from the given first-frame mask, and then associating object detections in future frames using Hungarian matching based on their bounding-box overlap. 

\PAR{Mask-guided.} Here, STCN consistently out-performs the box tracker for both class sets since it is a state-of-the-art method for exemplar-guided tracking whereas the box tracker is a basic approach. Note how the difference widens for uncommon classes where STCN achieves 49.2 \hotauncommon{} (validation) whereas box tracker only achieves 13.6. This is because STCN is class-agnostic, and can track any given first-frame object mask, whereas the box tracker uses object detections produced by a Mask2Former~\cite{Cheng21mask2former} model trained on COCO~\cite{Lin14coco} (\ie `common' classes).

\PAR{Box-guided.} For the box and point guided tasks, we formulate baselines by treating them as extensions of the mask-guided task with an additional `box $\rightarrow$ mask' or `point $\rightarrow$ mask' pre-processing step which regresses the segmentation mask from the given first-frame bounding-box or point, respectively.
For the box-guided task, we do this in two ways: (1) We compute the IoU between the given bounding box with all the bounding boxes of the image detections for that frame, and assign the mask belonging to the detection with the highest overlap, and (2) we input the given first-frame bounding box to a PointRend~\cite{Kirillov20pointrend} based mask regression head from a MaskRCNN~\cite{he17maskrcnn} model, and use the resulting segmentation mask. Looking at Table~\ref{tab:baselines}, we see that the scores for box-guided are generally lower than those for mask-guided due to the additional `box $\rightarrow$ mask' regression step. Among box-guided scores, PointRend performs much better than using the best-matched detection. For \hotauncommon{} in particular, the PointRend baseline achieves 44.3 (validation) whereas the matched detection baseline only gets 18.7. Given the fact that both networks (the image detector used for matching, and the PointRend mask head) are trained on COCO, it shows that PointRend is a much more robust `box $\rightarrow$ mask' regressor than matching detections.

\PAR{Point-guided.} Finally, for the point-guided task we implement `point $\rightarrow$ mask' by taking the mask for the highest scoring detection which contains the given point. Since this technique is prone to errors, we see that in terms of all three metrics, the baselines for the point-guided task are worse compared to those for box-guided and mask-guided. 

\subsection{Class-guided}

For the class-guided stream, we show two tracking-by-detection baselines for each task: (1) A simple box tracker which links per-frame object detections using box IoU followed by Hungarian matching, and (2) an `STCN tracker', where, in order to associate object detection masks in frame $t$ with those in $t+1$, we use STCN to propagate the masks from frame $t+1$ to frame $t$, and then use the IoU between these propagated masks and the object masks in frame $t$ as an association metric.

\PAR{Common.} This task requires segmenting and tracking objects belonging to the 78-class `common' set. Here, the STCN tracker performs better than the box tracker (51.2 vs. 45.5 \hotacommon{}) because STCN-based mask propagation is more accurate for temporal association compared to bounding-box IoU. 
%
%
By effectively utilizing STCN, a method designed to tackle the mask exemplar-guided task, for class-guided tracking, we exemplify the related nature of these tasks and the potential for knowledge exchange among them. Note that we do not evaluate the common task for `uncommon' classes since this is not required by the task definition.

\PAR{Long-tail.} Here, methods are required to segment and track objects belonging to the 482-class `all' set (see Sec.~\ref{sec:metrics} for class set definitions). We note that the scores here are significantly worse than those for the common task. This is because we used a MaskRCNN~\cite{he17maskrcnn} model trained on LVIS~\cite{Gupta19LVIS} to obtain object detections for this larger set of classes. We observed that the detections produced by this network are of poor quality. Even for the `common' classes, the performance of both baselines reduces drastically when these detections are used (27.0 \hotacommon{} on validation for box tracker vs. 45.5 for the common task). Also note that here, the STCN tracker performs worse than the box tracker, even though the opposite was true for the common task. The reason is that STCN performs erroneous mask propagation when the input mask quality is bad. Hence, in this case the more basic bounding-box IoU tracker performs comparatively better. To the best of our knowledge, this is the first time that a quantitative comparison of video object tracking methods is given in terms of their performance on such a large class set. We hope that our benchmark will encourage other researchers to discover ways of mitigating this large performance gap.

\PAR{Open-world.} Finally, for the open-world task, we use the $\mathrm{OWTA}$ metric which is similar to \hota{}, but without penalization for false positives. As per the task definition, methods can only be trained on the `common' class set, but during inference, are expected to additionally segment objects belonging to the `uncommon' set. Here, we again use the box tracker and STCN tracker with image-level detections from a Mask2Former~\cite{Cheng21mask2former} model trained on COCO. We additionally report results for the baseline proposed by Liu~\etal~\cite{liu22OWTB} (OWTB). Unsuprisingly, all methods suffer performance degradation for the `uncommon' set. The STCN tracker achieves the highest \hotaall{} score (64.6 on validation), but OWTB performs significantly better in terms of \hotauncommon{} (38.8) compared to the next best baseline (box tracker: 25.0).

\begin{figure}
\centering
\includegraphics[width=0.95\linewidth]{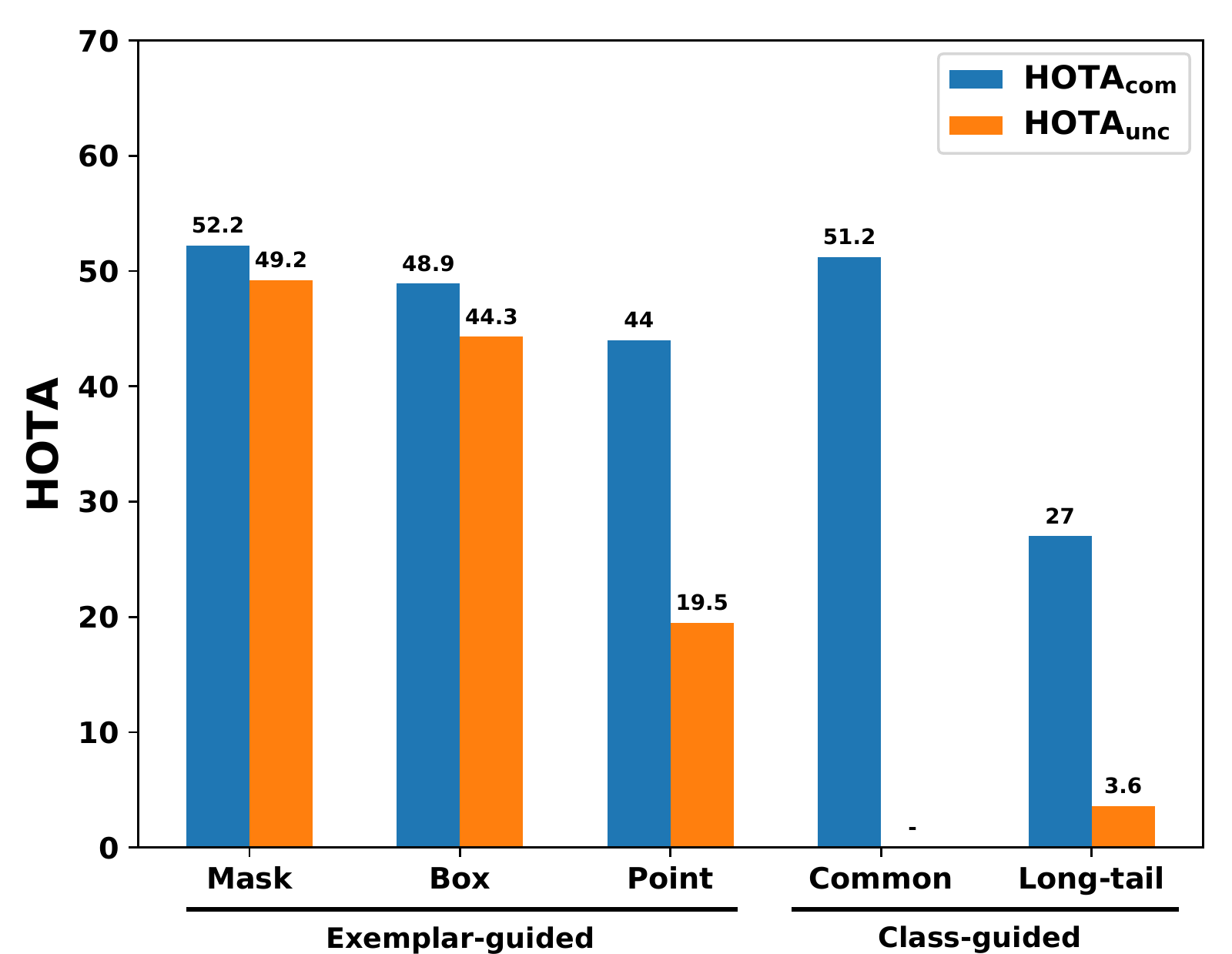}
\caption{Comparison of baseline performances for different tasks.}
\label{fig:combined_bar_chart}
\end{figure}

\subsection{Comparison Across Tasks}

Using consistent metrics enables us to directly compare results for different methods across different tasks. We illustrate this comparison for our baselines in Fig.~\ref{fig:combined_bar_chart} which charts the best-performing baseline on the validation set for each task. Note that we omitted open-world results since the $\mathrm{OWTA}$ metric differs slightly from \hota{}.
For \hotacommon{}, the mask exemplar-guided score of 52.2 is only slightly higher than that for the common class-guided task (51.2). 
This may seem surprising since the exemplar-guided task is inherently easier because, for each target object, methods have access to (1) an explicit cue (mask/box/point), and (2) the class label.
However, we noticed that exemplar-guided methods often lose the target object (\eg due to occlusion or erratic motion), and thereafter cannot recover it; for both STCN and box tracker, if the object is not track-able for more than a certain number of frames, it is assumed to have disappeared. On the other hand, class-guided methods predict an arbitrary number of tracks which may collectively cover more of the ground-truth object. Because \hota{} calculation involves Detection Accuracy ($\mathrm{DetA}$), the class-guided methods receive partial credit for correct per-frame detections even if the object ID is inconsistent/fragmented over time. This effect is more visible for \DatasetAbbrev{} because it contains longer videos ($\sim$30s) compared to existing exemplar-guided benchmarks~\cite{PontTuset17Davis,Xu18YouTubeVOS} ($\sim$5-10s).

In terms of \hotauncommon{} however, even the worst exemplar-guided score (19.5) is significantly higher than the 3.6 achieved for class-guided. This is because exemplar-guided methods are inherently class-agnostic and can propagate arbitrary object masks, but our class-guided tracking-by-detection baselines are very sensitive to per-frame object localization and classification quality, which is currently quite poor for this larger class set.


\section{Conclusion}
\label{sec:conclusion}

We present \DatasetAbbrev{}: a benchmark which unifies six tasks related to object recognition, segmentation and tracking in video with a clear task taxonomy and consistent evaluation metrics. Our dataset contains a large and diverse video set with pixel-precise masks for a large vocabulary of object classes. We temporally densified the object masks for the training set using a semi-automated pipeline which yields accurate results while drastically reducing human annotation effort. Finally, we presented a number of baselines for the proposed tasks and analyzed their performance. We hope that our benchmark will serve as a valuable resource for researchers to evaluate their object tracking methods.

\PAR{Acknowledgements.}
This project was partially funded by ERC Consolidator Grant DeeVise (ERC-2017-COG-773161) and the CMU Argo AI Center for Autonomous Vehicle Research. 
We would like to thank Berin Bala Chander and AnnotateX (\url{https://annotatex.com/}) for annotation support for this project.
Computing resources were granted by RWTH Aachen University under project `supp0003'. 
We thank Jonas Schult, Christian Schmidt, Alexey Nekrasov, Sabarinath Mahadevan and Markus Knoche for helpful discussions.

\clearpage

{\small
\bibliographystyle{ieee_fullname}
\bibliography{abbrev_short,references}
}

\clearpage
\twocolumn[{%
 \centering
 \LARGE \textbf{Supplementary Material\\[1cm]}
}]

\setcounter{equation}{0}
\setcounter{figure}{0}
\setcounter{table}{0}
\setcounter{page}{1}
\setcounter{section}{0}
\makeatletter
\renewcommand{\theequation}{S\arabic{equation}}
\renewcommand{\thefigure}{S\arabic{figure}}
\renewcommand{\thetable}{S\arabic{table}}
\renewcommand{\thesection}{S\arabic{section}}

\begin{table*}[h]

\centering{}
\setlength{\tabcolsep}{4.0pt}
\newcommand\RotText[1]{\rotatebox{90}{\parbox{2cm}{\centering#1}}}
\newcommand{\interTableSpace}[0]{\vspace{32pt}}
\footnotesize
\def\metricFontSize{\tiny}
\def\streamFontSize{\scriptsize}

\caption{
Extended baseline results for the validation set with multiple metrics.
Evaluation metrics are reported separately for `common', `uncommon' and `all' classes. 
Object detector training data: \textbf{*}: COCO, \textbf{$\dagger$}: LVIS. 
}
\label{tab:expanded_baselines_validation}
\vspace{4pt}

\newcolumntype{Y}{>{\centering\arraybackslash}X}
\setlength{\tabcolsep}{2pt}
\begin{tabularx}{\textwidth}{cclYYYcYp{5pt}YYYcYp{5pt}YYYcY}
\toprule 
 & & \multirow{2}{*}{Baseline Method} & \multicolumn{5}{c}{All} &  & \multicolumn{5}{c}{Common} &  & \multicolumn{5}{c}{Uncommon}\\
\cmidrule{4-8} \cmidrule{10-14} \cmidrule{16-20}
 &  &  & \metricFontSize{}$\mathrm{DetA}$ & \metricFontSize{}$\mathrm{AssA}$ & \metricFontSize{}$\mathrm{HOTA}$ & \metricFontSize{}$\mathrm{HOTA}^\text{obj}$ & \metricFontSize{} $\mathrm{mAP}$ &  & \metricFontSize{}$\mathrm{DetA}$ & \metricFontSize{}$\mathrm{AssA}$ & \metricFontSize{}$\mathrm{HOTA}$ & \metricFontSize{}$\mathrm{HOTA}^\text{obj}$ & \metricFontSize{} $\mathrm{mAP}$ &  & \metricFontSize{}$\mathrm{DetA}$ & \metricFontSize{}$\mathrm{AssA}$ & \metricFontSize{}$\mathrm{HOTA}$ & \metricFontSize{}$\mathrm{HOTA}^\text{obj}$ & \metricFontSize{} \metricFontSize{} $\mathrm{mAP}$\\
\cmidrule{1-8} \cmidrule{10-14} \cmidrule{16-20} 
\multirow{7}{10pt}{\centering{}\rotatebox{90}{\parbox{2cm}{\streamFontSize{}  \centering Exemplar-guided}}} & \multirow{2}{*}{\centering{}Mask} & %
STCN~\cite{Cheng21stcn}                    &  46.1    &   54.9    &   49.8    &   59.2    &   33.3    &   &   44.7    &   62.5    &   52.2    &   60.1    &   27.5    &   &   46.4    &   53.1    &   49.2    &   54.8    &   34.8\\
 &  & %
 Box Tracker{*}~\cite{luiten2020trackeval}         &  15.1    &   22.5    &   18.0    &   47.7    &   12.5    &   &   28.7   &    46.3    &   35.8    &   52.3    &   12.5    &   &   11.8    &   16.6    &   13.6   &    18.8    &   4.4\\
\cmidrule{2-8} \cmidrule{10-14} \cmidrule{16-20} 
 & \multirow{3}{*}{\centering{}Box} & %
STCN (PointRend~\cite{Kirillov20pointrend})        &  41.1    &   51.1    &   45.8    &   57.1    &   26.7    &   &   40.9    &   60.2    &   48.9    &   58.5    &   23.5    &   &   41.1    &   48.8    &   44.3    &   50.9    &   27.5\\
 &  & %
STCN (Matched Det{*})   &  20.0    &   32.4    &   24.5    &   54.4    &   13.3    &    &   38.7    &   60.8    &   47.6    &   57.6    &   22.7    &   &   15.4    &   25.3    &   18.7    &   34.2    &   11.0\\
 &  & Box Tracker{*} & 10.9 & 18.8 & 13.7 & 48.6 & 4.4 &  & 27.1 & 45.0 & 34.2 & 52.8 & 10.8 &  & 6.9 & 12.2 & 8.6 & 18.3 & 2.8\\
\cmidrule{2-8} \cmidrule{10-14} \cmidrule{16-20}
 & \multirow{2}{*}{\centering{}Point} & STCN (Matched Det{*}) & 19.9 & 32.1 & 24.4 & 48.4 & 12.8 &  & 34.4 & 59.0 & 44.0 & 51.6 & 20.6 &  & 16.3 & 25.4 & 19.4 & 31.4 & 10.8\\
 &  & Box Tracker{*} & 9.9 & 17.6 & 12.7 & 43.7 & 3.2 &  & 24.2 & 43.7 & 31.7 & 48.0 & 9.7 &  & 6.3 & 11.2 & 7.9 & 17.1 & 1.6\\
\cmidrule{1-8} \cmidrule{10-14} \cmidrule{16-20} 
\multirow{4}{10pt}{\centering{}\rotatebox{90}{\parbox{1.5cm}{\streamFontSize{} \centering Class-guided}}} & \multirow{2}{*}{\centering{}Common} & STCN Tracker{*} & - & - & - & - & - &  & 51.7 & 51.9 & 51.2 & 66.8 & 13.6 &  & - & - & - & - & -\\
 &  & Box Tracker{*} &  &  &  &  &  &  & 51.7 & 43.0 & 45.5 & 62.9 & 9.9 &  & - & - & - & - & -\\
\cmidrule{2-8} \cmidrule{10-14} \cmidrule{16-20}
 & \multirow{2}{*}{\centering{}Long-Tail} & STCN Tracker$^{\dagger}$ & 8.0 & 13.4 & 5.5 & 13.5 & 0.9 &  & 27.0 & 13.4 & 17.5 & 14.4 & 0.7 &  & 3.2 & 2.4 & 2.5 & 8.9 & 0.6\\
 &  & Box Tracker$^{\dagger}$ & 8.3 & 29.5 & 8.2 & 18.6 & 1.4 &  & 28.0 & 29.5 & 27.0 & 19.8 & 3.0 &  & 3.4 & 4.3 & 3.6 & 11.7 & 0.9\\
\bottomrule
\end{tabularx}


\interTableSpace{}
\centering{}
\setlength{\tabcolsep}{4.0pt}
\footnotesize

\caption{
Extended baseline results for the test set with multiple metrics.
Evaluation metrics are reported separately for `common', `uncommon' and `all' classes. 
Object detector training data: \textbf{*}: COCO, \textbf{$\dagger$}: LVIS. 
}
\label{tab:expanded_baselines_test}
\vspace{4pt}

\setlength{\tabcolsep}{2pt}
\begin{tabularx}{\textwidth}{cclYYYcYp{5pt}YYYcYp{5pt}YYYcY}
\toprule 
 & & \multirow{2}{*}{Baseline Method} & \multicolumn{5}{c}{All} &  & \multicolumn{5}{c}{Common} &  & \multicolumn{5}{c}{Uncommon}\\
\cmidrule{4-8} \cmidrule{10-14} \cmidrule{16-20}
 &  &  & \metricFontSize{}$\mathrm{DetA}$ & \metricFontSize{}$\mathrm{AssA}$ & \metricFontSize{}$\mathrm{HOTA}$ & \metricFontSize{}$\mathrm{HOTA}^\text{obj}$ & \metricFontSize{} mAP &  & \metricFontSize{}$\mathrm{DetA}$ & \metricFontSize{}$\mathrm{AssA}$ & \metricFontSize{}$\mathrm{HOTA}$ & \metricFontSize{}$\mathrm{HOTA}^\text{obj}$ & \metricFontSize{} mAP &  & \metricFontSize{}$\mathrm{DetA}$ & \metricFontSize{}$\mathrm{AssA}$ & \metricFontSize{}$\mathrm{HOTA}$ & \metricFontSize{}$\mathrm{HOTA}^\text{obj}$ & \metricFontSize{} \metricFontSize{} mAP\\
\cmidrule{1-8} \cmidrule{10-14} \cmidrule{16-20} 
\multirow{7}{10pt}{\centering{}\rotatebox{90}{\parbox{2cm}{\streamFontSize{}  \centering Exemplar-guided}}} & \multirow{2}{*}{\centering{}Mask} & %
STCN~\cite{Cheng21stcn} & 48.8 & 57.3 & 52.4 & 60.7 & 34.4 &  & 45.0 & 59.2 & 51.1 & 61.0 & 30.0 &  & 49.6 & 57.0 & 52.7 & 59.5 & 35.3\\
 &  & Box Tracker{*}~\cite{luiten2020trackeval} & 12.1 & 17.2 & 14.1 & 44.4 & 3.5 &  & 21.9 & 37.3 & 28.0 & 48.5 & 9.2 &  & 10.2 & 13.2 & 11.4 & 16.9 & 2.4\\
\cmidrule{2-8} \cmidrule{10-14} \cmidrule{16-20} 
 & \multirow{3}{*}{\centering{}Box} & STCN (Pointrend~\cite{Kirillov20pointrend}) & 41.9 & 51.7 & 46.0 & 58.1 & 25.6 &  & 42.2 & 58.0 & 48.9 & 59.0 & 26.2 &  & 41.9 & 50.5 & 45.4 & 53.6 & 25.4\\
 &  & STCN (Matched Det{*}) & 20.6 & 32.9 & 25.0 & 53.9 & 13.6 &  & 33.1 & 55.5 & 41.9 & 56.6 & 20.5 &  & 18.1 & 28.4 & 21.7 & 36.6 & 12.2\\
 &  & Box Tracker{*} & 11.6 & 16.9 & 13.6 & 44.2 & 3.0 &  & 21.5 & 37.2 & 27.7 & 48.3 & 8.3 &  & 9.6 & 12.8 & 10.8 & 17.0 & 1.9\\
\cmidrule{2-8} \cmidrule{10-14} \cmidrule{16-20}
 & \multirow{2}{*}{\centering{}Point} & STCN (Matched Det{*}) & 20.6 & 32.5 & 24.9 & 49.6 & 12.5 &  & 30.8 & 53.8 & 39.5 & 52.3 & 17.7 &  & 18.6 & 28.3 & 22.0 & 34.6 & 11.5\\
 &  & Box Tracker{*} & 8.0 & 14.1 & 10.1 & 41.6 & 2.5 &  & 17.8 & 36.1 & 24.4 & 45.5 & 6.7 &  & 6.1 & 9.7 & 7.3 & 16.1 & 1.7\\
\midrule 
\multirow{4}{10pt}{\centering{}\rotatebox{90}{\parbox{1.5cm}{\streamFontSize{} \centering Class-guided}}} & \multirow{2}{*}{\centering{}Common} & STCN Tracker{*} & - & - & - & - & - &  & 37.1 & 34.9 & 34.6 & 58.8 & 6.8 &  & - & - & - & - & -\\
 &  & Box Tracker{*} & - & - & - & - & - &  & 37.1 & 35.5 & 34.3 & 56.9 & 6.8 &  & - & - & - & - & -\\
\cmidrule{2-8} \cmidrule{10-14} \cmidrule{16-20}
 & \multirow{2}{*}{\centering{}Long-Tail} & STCN Tracker$^{\dagger}$ & 5.4 & 4.6 & 4.5 & 13.4 & 0.3 &  & 19.6 & 16.7 & 17.1 & 14.3 & 1.7 &  & 2.6 & 2.2 & 2.0 & 8.0 & 0.0\\
 &  & Box Tracker$^{\dagger}$ & 5.4 & 6.8 & 5.7 & 15.4 & 0.3 &  & 19.6 & 23.2 & 20.1 & 16.3 & 1.7 &  & 2.6 & 3.6 & 2.9 & 10.7 & 0.0\\
\bottomrule
\end{tabularx}


\interTableSpace{}
\caption{Extended baseline results for the open-world class-guided task for the validation set.}
\label{tab:extended_openworld_baselines_validation}
\centering
\footnotesize
\vspace{4pt}
\setlength{\tabcolsep}{2pt}
\begin{tabularx}{0.8\linewidth}{lp{3pt}YYYp{3pt}YYYp{3pt}YYY}
\toprule
\multirow{2}{*}{Baseline Method} & & \multicolumn{3}{c}{All} & & \multicolumn{3}{c}{Common} & & \multicolumn{3}{c}{Uncommon} \\
\cmidrule{3-5} \cmidrule{7-9} \cmidrule{11-13}
                & & $\mathrm{DetRe}$ & $\mathrm{AssA}$ & $\mathrm{OWTA}$ & & $\mathrm{DetRe}$ & $\mathrm{AssA}$ & $\mathrm{OWTA}$ & & $\mathrm{DetRe}$ & $\mathrm{AssA}$ & $\mathrm{OWTA}$ \\
\cmidrule{1-1} \cmidrule{3-5} \cmidrule{7-9} \cmidrule{11-13}
STCN Tracker           & & 67.0 & 62.6  & 64.6  & & 78.8 & 64.1  & 71.0  & & 20.0  & 33.3  & 25.0  \\
Box Tracker            & & 66.9 & 55.8  & 60.9  & & 78.7 & 57.1  & 60.9  & & 20.1  & 30.5  & 24.0  \\
OWTB~\cite{liu22OWTB}  & & 70.9 & 45.2  & 56.2  & & 76.8 & 47.0  & 59.8  & & 46.5  & 34.3  & 38.5  \\
\bottomrule
\end{tabularx}


\interTableSpace{}
\caption{Extended baseline results for the open-world class-guided task for the test set.}
\label{tab:extended_openworld_baselines_test}
\centering
\footnotesize
\vspace{4pt}
\setlength{\tabcolsep}{2pt}
\begin{tabularx}{0.8\linewidth}{lp{3pt}YYYp{3pt}YYYp{3pt}YYY}
\toprule
\multirow{2}{*}{Baseline Method} & & \multicolumn{3}{c}{All} & & \multicolumn{3}{c}{Common} & & \multicolumn{3}{c}{Uncommon} \\
\cmidrule{3-5} \cmidrule{7-9} \cmidrule{11-13}
                & & $\mathrm{DetRe}$ & $\mathrm{AssA}$ & $\mathrm{OWTA}$ & & $\mathrm{DetRe}$ & $\mathrm{AssA}$ & $\mathrm{OWTA}$ & & $\mathrm{DetRe}$ & $\mathrm{AssA}$ & $\mathrm{OWTA}$ \\
\cmidrule{1-1} \cmidrule{3-5} \cmidrule{7-9} \cmidrule{11-13}
STCN Tracker           & & 61.6 & 54.1  & 57.5  & & 71.5  & 55.7  & 62.9 & & 21.0  & 28.6  & 23.9  \\
Box Tracker            & & 61.5 & 51.1  & 55.9  & & 71.4  & 52.5  & 61.0 & & 21.1  & 30.0  & 24.6  \\
OWTB~\cite{liu22OWTB}  & & 70.7 & 45.5  & 56.3  & & 76.6  & 47.3  & 59.9 & & 45.7  & 33.6  & 38.3  \\
\bottomrule
\end{tabularx}

\end{table*}

\section{Dataset Visualization}

\begin{figure*}
    \centering
    \includegraphics[width=\textwidth,height=0.98\textheight,keepaspectratio]{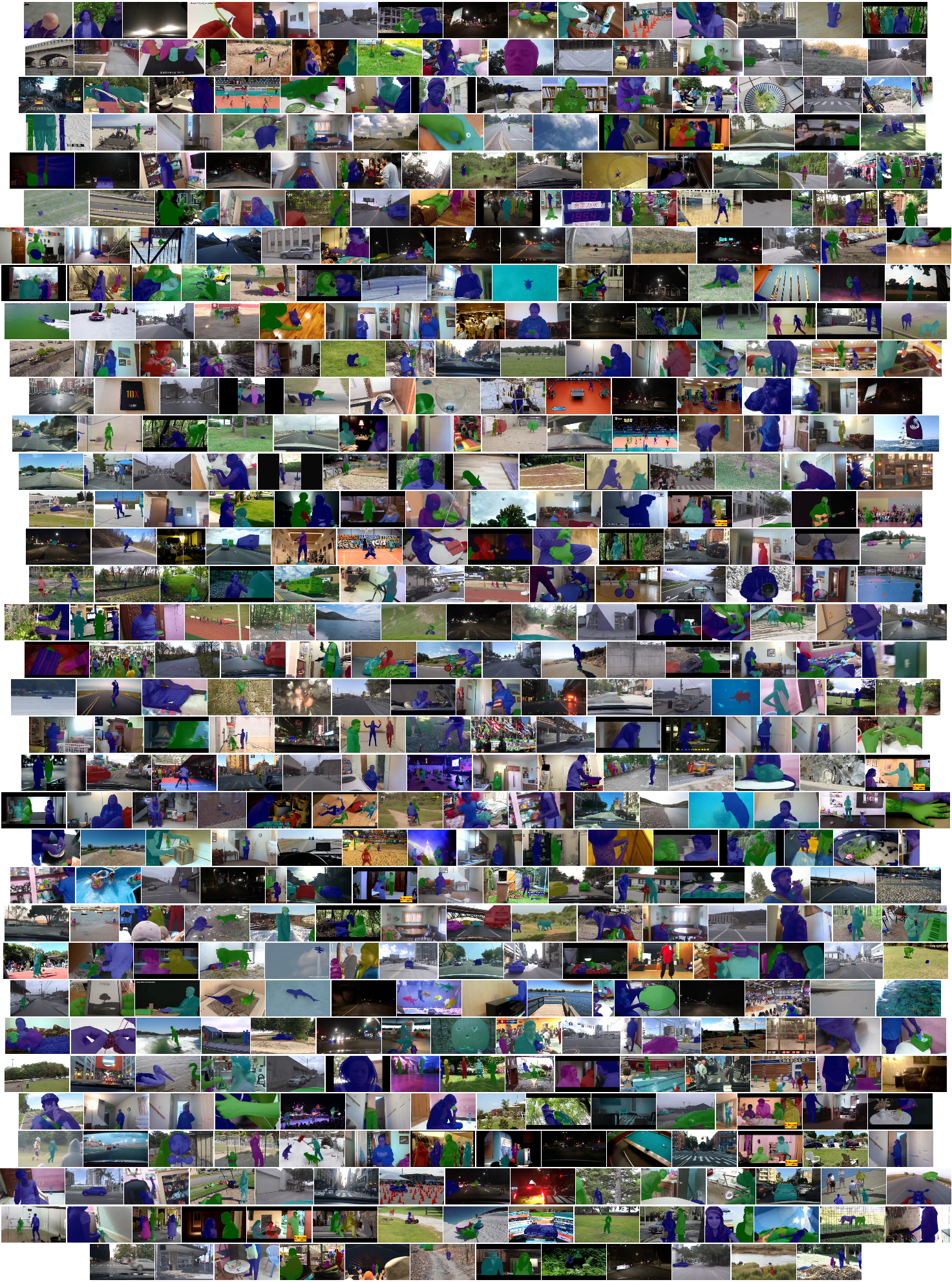}
    \caption{Screenshot of the dataset viewer supplied in the supplemental.zip. Here assorted examples of annotated images from the \DatasetAbbrev{} training set are shown, however in the dataset viewer full videos with tracked mask annotations are shown.}
    \label{fig:annotation_tile}
\end{figure*}

\ifArxivMode
Examples of some video frames with annotated masks from our training set are shown in Fig.~\ref{fig:annotation_tile}.
\else
The supplementary zip file contains a dataset viewer, for viewing all of the videos and annotations in the \DatasetAbbrev{} dataset. Please refer to the `readme' file for further instructions. A screenshot from this viewer, showing examples of some video frames with annotated masks from our training set is shown in Fig.~\ref{fig:annotation_tile}.
\fi

\section{Extended Baseline Results}

Tables~\ref{tab:expanded_baselines_validation} and~\ref{tab:expanded_baselines_test} show extended results for our baselines (Sec.~7 in main text) for the validation and test set, respectively. Tables~\ref{tab:extended_openworld_baselines_validation} and~\ref{tab:extended_openworld_baselines_test} show extended results for the open-world class-guided task for the validation and test sets, respectively. Here, aside from the \hota{} score, we also provide the \detA{}, \assA{} and mAP scores. Additionally, we provide another variant of \hota{} called \hotaobj{} where the final score is calculated by averaging the per-object \hota{} scores, instead of averaging over the object classes. We draw the following observations and comparisons from the tabulated results:

\subsection{HOTA vs. mAP}
mAP (mean average precision) is used as a metric by several existing benchmarks related to video object tracking and segmentation~\cite{Yang19YouTubeVIS,Qi21OVIS,Dave20Tao}. It works by computing the mask IoU at the track-level (\ie across the whole video) and then uses Hungarian matching to assign at most one predicted track to each ground-truth track. All other predicted tracks are considered as false positives, even if they intersect strongly with a ground-truth track. On the other hand, \hota{} gives weighting to both per-frame detection accuracy (\detA{}) as well as temporal association accuracy (\assA{}). 

This difference can be noted by comparing the mask exemplar-guided results with those for the common class-guided task. We see in Table~\ref{tab:expanded_baselines_validation} that in terms of \hota{} on common classes, STCN achieves 52.2 which is only slightly higher than the 51.2 achieved by the STCN Tracker. However, in terms of mAP, the difference is much larger (27.5 vs. 13.6). 
This is because the mask exemplar-guided task provides the first-frame ground-truth mask for every object during inference. This often results in one predicted track having reasonably good overlap with the ground-truth.
However, once the target object is lost due to tracking errors, it can no longer be recovered. This results in a higher mAP score because there is usually one good match in the predictions for each ground-truth track, and no false positives. For the class-guided task however, the first-frame mask is not given, and methods predict an arbitrary number of tracks which may collectively capture a certain ground-truth object (in multi-object tracking parlance, we would say that the predicted tracks are \emph{fragmented}). This results in a low mAP score because there is no single high-quality predicted track which has a high IoU with the ground-truth, and every predicted track aside from the best-matched one is considered as a false positive. 

With \hota{} however, we can quantitatively analyze this phenomenon (Table~\ref{tab:expanded_baselines_validation}): in terms of \detA{}, the exemplar-guided STCN achieves 44.7 which is lower than the 51.7 for the class-guided STCN Tracker. This shows that in terms of per-frame detections, the class-guided method correctly predicts more of the ground-truth. In terms of \assA{} however, STCN achieves 62.5 which is higher than the 51.9 for the STCN Tracker. This shows that temporal association quality is better for the exemplar-guided method (as we hypothesized earlier).

Finally, we note that some methods have an mAP score of 0.0 for some settings \eg uncommon class set for the long-tail class-guided task (Table~\ref{tab:expanded_baselines_validation}). This happens when the predicted object tracks cannot be associated with any ground-truth track because their IoUs are all below the acceptance threshold. Note that mAP is computed as an average over several different threshold values, so an overall mAP of 0.0 implies that even the lowest threshold was not satisified by any predicted track in the entire dataset. This further highlights the potential for improvement for the long-tail task, both in terms of per-frame object detections as well as temporal association.

\subsection{DetA vs. AssA}
The ability to numerically quantify per-frame detection and temporal association quality can prove useful in analyzing strengths and weaknesses of various methods. For instance, when comparing STCN Tracker and Box Tracker for class-guided tasks, we note that the difference in the final \hota{} scores arises mainly from the the difference in \assA{}. This is understandable because both methods use the same set of per-image detection masks. Any small differences in \detA{} arise from the fact that we perform a post-processing step where very short object tracks (\eg those containing just one detection) are discarded.

\subsection{$\mathbf{HOTA}$ vs. $\mathbf{HOTA}^\textbf{obj}$}
In general, we note that the \hotaobj{} scores are higher than \hota{}. We recap that the former gives equal weight to each object track when computing the final score, whereas the latter gives equal weight to each object class, even if the number of ground-truth tracks in the object classes are unbalanced. As a result, the \hota{} score is pushed down by the poor performance of the method on a few object classes even though they contain only a few object tracks.

\subsection{Comparing Open-world to Other Tasks}

In the main text, we discussed how the open-world class-guided task uses $\mathrm{OWTA}$ as an evaluation metric, which is a modified version of \hota{} where \detA{} is replaced with \detRe{} (Detection Recall). In other words, \detRe{} is a modification of \detA{} in which false positives are not penalized.

To quantitatively compare the open-world result to other tasks, we can analyze the difference in \detA{} and \detRe{} for the same class split and baseline method. We see that the \detRe{} is consistently higher than \detA{}: for STCN Tracker on the validation set, \detRe{} is 78.8 for the `common' class set (Table~\ref{tab:extended_openworld_baselines_validation}) compared to a \detA{} of 51.7 for the common class-guided task (Table~\ref{tab:expanded_baselines_validation}). The numerical difference between these two arises due to the presence of false positives in the image-level object detector output.

Secondly, we point out that the metric \owta{} can be seen as an upper-bound for the \hotaobj{} metric which neglects penalization for false positives. Hence, the 52.3 \hotaobj{} score achieved by STCN tracker for the common class-guided task (Table~\ref{tab:expanded_baselines_validation}) is lower than the 71.0 \owta{} achieved by STCN Tracker on common classes for the open-world task (Table~\ref{tab:extended_openworld_baselines_validation}). This difference is again attributable to the presence of false positive detections in the method's predicted output.

\section{Miscellaneous Implementation Details}

\PAR{Point Selection Criterion for Exemplar-guided Task.}
For the point exemplar-guided task, we select the point coordinate to provide to the method as follows: we compute the distance between all points inside an object mask to the closest point on the object boundary and choose the point with the highest distance, \ie the `inner-most' point of the object mask. If multiple points share the highest distance value, we select the point closest to the centroid of the object mask.

\PAR{Image-level Object Detector.}
As mentioned in the main text, for obtaining image-level object masks for common classes, we employ a Mask2Former~\cite{Cheng21mask2former} network trained on COCO~\cite{Lin14coco}. In particular, we use the best-performing model checkpoint provided by the authors which uses a Swin-L~\cite{liu2021Swin} backbone. For the long-tail task, we use the best-performing Mask-RCNN~\cite{he17maskrcnn} model provided by Detectron2~\cite{wu2019detectron2} which is trained on LVIS~\cite{Gupta19LVIS} and has a ResNeXt-101 backbone.

\end{document}